\documentclass[journal]{IEEEtran} 
\IEEEoverridecommandlockouts

\usepackage{multirow}
\usepackage{tabularx, booktabs, adjustbox}
\usepackage{balance}

\usepackage{siunitx} 
\usepackage{amssymb}

\usepackage{epsfig}
\usepackage{float}
\usepackage{color}
\usepackage{url}

\usepackage{algorithm}      
\usepackage{algorithmicx}   
\usepackage{algpseudocode}  

\algdef{SE}[DOWHILE]{Do}{doWhile}{\algorithmicdo}[1]{\algorithmicwhile\ #1}%

\DeclareMathAlphabet\mathbfcal{OMS}{cmsy}{b}{n} 

\usepackage{cite}
\usepackage{amssymb, amsfonts, bbold}
\usepackage{amsmath}
\usepackage{mathtools}
\usepackage{bbm}

\usepackage{caption}
\usepackage{subcaption}

\usepackage{bm}

\usepackage{amsthm}

\let\labelindent\relax
\usepackage{enumitem}        
\setenumerate[enumerate]{align=left}

\usepackage{multirow}
\usepackage{physics}

\usepackage{cite}
\makeatletter
\let\NAT@parse\undefined
\makeatother
\usepackage{hyperref}  

\newcommand{\ul}[1]{\underline{#1}} 
\newcommand{\tb}[1]{\textbf{#1}} 

\newcommand{\floor}[1]{\left\lfloor #1 \right\rfloor}
\newcommand{\sksym}[1]{\floor{#1}_{\times}}
\newcommand{\sparse}[1]{\operatorname{sparse}(#1)}

\newcommand{\expectation}[1]{\mathbb{E}\left[#1\right]}
\newcommand{\est}[1]{\hat{#1}}

\hypersetup{
	colorlinks=true,
	linkcolor=blue,
	filecolor=blue,      
	urlcolor=blue,
	citecolor=blue,
}

\def\R{\mathbb{R}}

\def\WorldFrame{\{\mathcal{W}\}}
\def\VOFrame{\{\mathcal{V}\}}
\def\CameraFrame{\{\mathcal{C}\}}

\def\NumData{k}
\def\StateVector{\mathbf{\Theta}}
\def\SlidingWindow{\mathcal{K}}

\def\UWBAnchorn{a_n}
\def\UWBAntenna{\mathfrak{a}}

\def\CovMatR{\mathbf{\Sigma}}
\def\CovMatS{\mathcal{S}}

\def\ParamGAT{\mathcal{A}}
\def\ParamScale{s}
\def\ParamRotMat{\mathbf{R}}
\def\ParamRasVec{\mathbf{r}}
\def\ParamTrans{\mathbf{t}}

\def\ParamRotVecxyz{\bar{v}}
\def\ParamRotAngle{\theta}

\def\ParamRotVec{\mathbf{v}}
\def\ParamRotAxis{\bar{\ParamRotVec}}
\def\VEdge{\bm{\gamma}}

\def\SolVector{\mathbf{c}}
\def\DerivativeMat{\mathbf{B}}

\def\PartialFv{\bm{\Phi}}
\def\PartialFs{\mu}
\def\PartialRv{\bm{\Gamma}}

\def\VCxyz{o}
\def\VCp{\mathbf{o}}
\def\Wap{\bm{a}_n}
\def\Cap{\bm{\rho}}
\def\UnitVec{\mathbf{u}}
\def\WCp{\prescript{\mathcal{W}}{\mathcal{C}}{\mathbf{p}}}
\def\WCq{\prescript{\mathcal{W}}{\mathcal{C}}{\mathbf{q}}}

\def\JacobianMat{\mathbf{J}}
\def\DetSubTrans{\alpha}
\def\DetSubRot{\beta}

\def\IdMat{\mathbf{I}}
\def\AlignedPos{\ParamRotMat \VCp}
\def\BodyVec{\mathbf{b}}

\def\StateVector{\mathbf{\Theta}}
\def\FIM{\mathbf{F}}
\def\SubFIM{\mathbf{\Lambda}}
\def\detF{\det(\FIM)}
\def\detL{\det(\SubFIM)}

\def\CovDR{\mathbf{\Sigma}}

\def\VarR{\sigma_r}

\def\UnitX{\UnitVec_x}
\def\UnitY{\UnitVec_y}
\def\UnitZ{\UnitVec_z}

\def\RangeKNtrue{d_\NumData}
\def\RangeKNnoisy{\tilde{d}_\NumData}
\def\RangeKNoise{\eta_\NumData}

\DeclareMathOperator*{\argmin}{arg\,min}

\def\QCQPStateVec{\mathbf{x}}
\def\QCQPLeftMat{\mathbf{D}}
\def\QCQPConsMatrix{\mathbf{P}}
\def\QCQPConsScalar{p}

\allowdisplaybreaks

\title{VR-SLAM: A Visual-Range Simultaneous Localization and Mapping System using Monocular Camera and Ultra-wideband Sensors}

\author{Thien Hoang Nguyen,
        Shenghai Yuan,
        and Lihua Xie,~\IEEEmembership{Fellow,~IEEE}
        
\thanks{This work is supported by the National Research Foundation, Singapore under its Medium Sized Center for Advanced Robotics Technology Innovation. (Corresponding author: Lihua Xie).\\
The authors are with School of Electrical and Electronic Engineering, Nanyang Technological University, Singapore 639798, 50 Nanyang Avenue (e-mail: e180071@e.ntu.edu.sg, shyuan@ntu.edu.sg, elhxie@ntu.edu.sg).}
}

\usepackage[para]{footmisc} 

\begin{document}

\bstctlcite{BSTcontrol}

\maketitle

\begin{abstract}
In this work, we propose a simultaneous localization and mapping (SLAM) system using a monocular camera and Ultra-wideband (UWB) sensors. Our system, referred to as VR-SLAM, is a multi-stage framework that leverages the strengths and compensates for the weaknesses of each sensor. Firstly, we introduce a UWB-aided 7 degree-of-freedom (scale factor, 3D position, and 3D orientation) global alignment module to initialize the visual odometry (VO) system in the world frame defined by the UWB anchors. This module loosely fuses up-to-scale VO and ranging data using either a quadratically constrained quadratic programming (QCQP) or nonlinear least squares (NLS) algorithm based on whether a good initial guess is available.
Secondly, we provide an accompanied theoretical analysis that includes the derivation and interpretation of the Fisher Information Matrix (FIM) and its determinant.
Thirdly, we present UWB-aided bundle adjustment (UBA) and UWB-aided pose graph optimization (UPGO) modules to improve short-term odometry accuracy, reduce long-term drift as well as correct any alignment and scale errors.
Extensive simulations and experiments show that our solution outperforms UWB/camera-only and previous approaches, can quickly recover from tracking failure without relying on visual relocalization, and can effortlessly obtain a global map even if there are no loop closures.
\end{abstract}


\section{Introduction} \label{sec:intro}

A fundamental challenge for autonomous mobile robots is achieving accurate, reliable and consistent localization in both the short and long terms. To this end, combining different sensor modalities, e.g. camera, LiDAR, radar, IMU and GPS, for their complementary advantages is a popular approach \cite{huang2019visual}. Extensive researches on this topic have shown impressive performance in various scenarios \cite{shenghai2021ussurvey}. However, since each sensor modality has its own benefits and drawbacks, choosing the right sensors for the environment while also satisfying the size, weight, power and cost constraints can be a challenging task. For example, LiDAR-based SLAM systems can produce very accurate odometry and map of the environment, but current commercial sensors are still very expensive, heavy and power hungry. GPS can be useful in outdoor environments but very unreliable in indoors, underground, cave or tunnel etc. Camera can provide rich information about the environment, but is susceptible to unfavorable visual conditions (extreme illumination, dynamic lighting, lack of distinct features etc.)

\begin{figure}[t]
\centering
	\includegraphics[width=\linewidth]{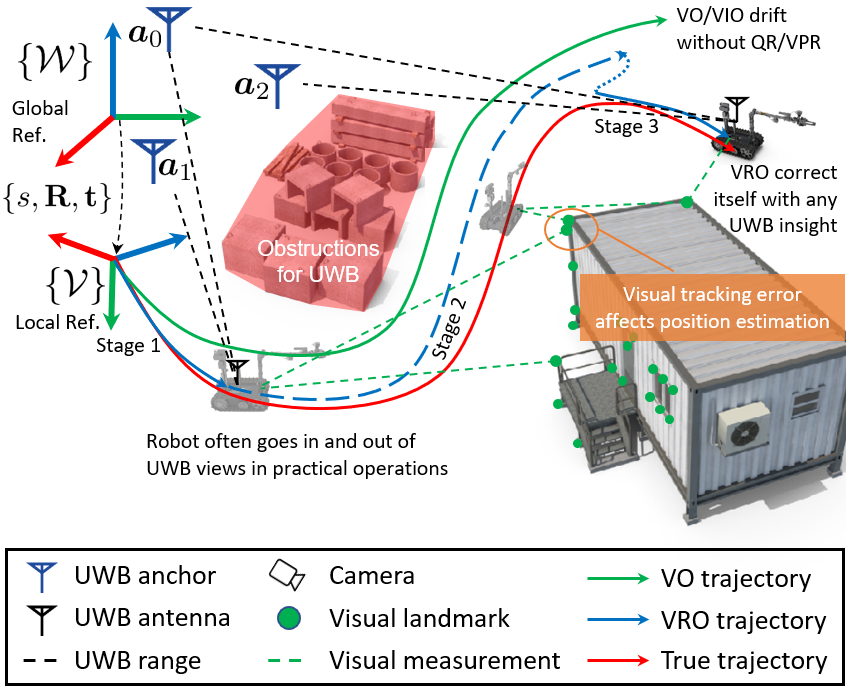}
    \caption{Overview of VR-SLAM, which operates in one of three stages: 1) Loosely-coupled initialization, 2) Tightly-coupled visual-range odometry, 3) UWB-aided relocalization.}
    \label{fig:overview_operation}
\end{figure}

In this context, UWB sensor provides several key advantages \cite{shule2020mulituwbsurvey}: different from VO, UWB distance measurements are drift-free and unaffected by visual conditions; unlike GPS, UWB can be used both indoors and outdoors; compared to LiDAR, UWB sensor is smaller, lighter, more affordable and simpler to install on a robot platform. UWB can also be employed in multi-robot cases as a communication network \cite{guo2016ultra,ding2021survey}, as a variable baseline between two robots \cite{karrer2021distributed}, or as inter-robot constraints for pose graph optimization \cite{boroson2020uwbmultiposegraph} or formation control \cite{nguyen2019distance}. On the other hand, UWB sensor requires good line-of-sight (LoS) for accurate measurement \cite{gonzalez2022uwb}, is unable to provide any perceptual information about the environment, and the performance of range-based localization methods relies on having enough UWB anchors installed and their configuration does not fall into a degenerate case \cite{shenghai2021ussurvey}.
As such, using UWB to complement other sensor modalities has great potential for real-life applications. Many researches have been put forth to combine UWB with RGB-D camera \cite{perez2017multi}, camera and GPS \cite{Lee2020ScaleUWB}, IMU \cite{li2018accurate}, IMU and camera \cite{wang2017ultra,shenhan2022viro} or LiDAR \cite{song2019uwb} and a combination of them \cite{nguyen2022viralfusion}.
Among these options, UWB and camera are not only low-cost and low-power but also mechanically flexible to install on robotic platforms. Hence, integrating UWB and camera sensors is a promising direction for real applications. 

To this end, a few challenges arise, which are illustrated in Fig. \ref{fig:overview_operation}.
First is finding the 3D affine transformation matrix that can scale, rotate and translate the VO system's data (including reference frame, keyframes and map points) into the world frame. We refer to this as the global affine transformation (GAT) problem.
Second, since the VO's estimate will drift over time while UWB range will not, we need to integrate visual and ranging measurements in a tightly-coupled manner to reduce the accumulated errors and improve the overall accuracy, which will be referred to as the visual-range odometry (VRO) problem. 
Third, while loop closure is a useful tool to correct the long-term drift or relocalize in case the tracking frontend fails (e.g., pure rotation motion, extreme brightness/darkness illumination, motion blur, etc.), it is susceptible to perceptual aliasing issues (i.e., different places have similar appearances). In our method, we combine both visual and ranging data to avoid admitting wrong loop closure candidates, accurately relocalize when tracking is lost, or at least re-initialize the system in the world frame as fast as possible. We refer to this as the UWB-aided visual relocalization (UVR) problem.

In this work, we propose a visual-range SLAM system, named VR-SLAM, which fuses a monocular camera and UWB sensors to produce metric-scaled, drift-free, and globally consistent odometry and mapping results. 
Our system is designed to address the aforementioned challenges (GAT, VRO, UVR) by appropriately fusing the measurements in a loosely or tightly-coupled manner depending on the situation.
The key advantages of our solution are: 1) we can achieve more extensive, accurate, and reliable results than either camera or UWB sensor alone, 2) we can avoid spurious loop closure candidates and accurately recover from a tracking failure, even if visual relocalization is not possible and 3) we can merge the maps from multiple operations even if there is no loop-closure between the individual maps. Our main contributions include:

\begin{itemize}

\item A theoretical analysis of the GAT problem, which focuses on the derivation of the FIM, interpretation of singular configurations and finding the uncertainty;

\item Optimization-based approaches, namely QCQP for initialization and NLS for refinement, for the GAT problem, which take advantage of the previous theoretical analysis;

\item A multi-stage SLAM framework that appropriately fuses camera and UWB sensors in different manners to solve the GAT, VRO and UVR problems as the need arises;

\item A thorough evaluation process including simulation, experiments with public datasets as well as real-life robot in large-scale, challenging conditions.

\end{itemize}

The rest of the paper is structured as follows. First, we review the main related works in Sect. \ref{sec:lit_review}. Then, in Sect. \ref{sec:ProbFormulation}, we outline the notations as well as the problem formulation. Next, the theoretical analysis for the GAT problem is presented in Sect. \ref{sec:TheorecticalAnalysis}. In Sect. \ref{sec:ProposedSystem}, our VR-SLAM is presented in details. Then, Sect. \ref{sec:ExpResults} presents the evaluation of VR-SLAM in simulations as well as experiments with the NTU-VIRAL dataset \cite{nguyen2021viraldataset}. Sect. \ref{sec:Conclusions} then concludes our work.

\section{Literature Review} \label{sec:lit_review}

\begin{table}[t]
\begin{adjustbox}{width=\columnwidth}
    \begin{tabular}[t]{c|c|c|c}
    \toprule
        \multirow{2}{*}{Method} & \multicolumn{3}{c}{Estimation problem} \\
        \cline{2-4}
        & GAT & VRO & UVR \\
    \hline \hline
        \cite{zhu2017planar} & scale + 2D trans. & - & - \\ 
    \hline
        \cite{shariati2016recovering} & scale + 2D trans. + 1D rot. & - & - \\ 
    \hline
        \cite{tiemann2018enhanced} & scale + 3D trans. + 1D rot. & - & - \\ 
    \hline \hline
        \cite{Thien2022scaleSDP} & scale + 3D anchor position & - & - \\ 
    \hline
        \cite{Shi2018UWBUnknown} & scale + 3D anchor position & UBA & - \\ 
    \hline
        \cite{Thien2020AuRO} & scale + 3D anchor position & UBA & - \\
    \hline \hline
        \textbf{Ours} & scale + 3D trans. + 3D rot. & UBA, UPGO & \checkmark \\
    \bottomrule
    \end{tabular}
\end{adjustbox}
\caption{Related works on GAT, VRO and UVR problems.}
\label{table:lit_review}
\end{table}

Table \ref{table:lit_review} summarizes the main related works on single-robot localization based on camera and UWB sensors.
On the one hand, the solutions for 2D GAT problem have been proposed given the UWB anchor's position and robot's initial heading \cite{zhu2017planar} or a large enough number of random samples \cite{shariati2016recovering} or that the robot always performs a pre-defined trajectory inside a full UWB localization system before the estimation \cite{tiemann2018enhanced}. However, these methods only tackle simplified versions of the GAT problem for ground robots and would not work for aerial robots. 
Furthermore, they only perform the estimation once and do not fine-tune the estimates over time, which we believe is a pivotal step if the results are to be used for VRO.
As such, these systems also do not address the VRO problem, which means the odometry will inevitably drift over time.

On the other hand, the VRO problem with any number of UWB anchors has been investigated in \cite{Shi2018UWBUnknown,Thien2020AuRO,Thien2022scaleSDP}. Instead of transforming the VO's system into the world frame, these methods find the anchor's position in the VO's reference frame, fix the estimates then combine visual and ranging data in a single optimization. By doing so, the initial estimation problem becomes similar to range-based target localization, which has been studied extensively \cite{guo2017reloc,nguyen2019single}. However, the map and anchor positions will be in a separate local frame in every run, which means the results are not globally consistent. 
Overall, existing solutions only tackle either the (simplified) GAT or VRO problems, while the UVR problem is ignored. Furthermore, most approaches assume the robot's trajectory will be sufficient to avoid degenerate configurations but fail to provide any theoretical analysis or method to detect singular cases or calculate the uncertainty of the estimate.
To the best of our knowledge, our system is the first to handle all of these problems in a single framework and in the general case.

Some recent works have studied the topic of range-aided pose graph optimization for single \cite{funabiki2021uwbmultiposegraph} and multi-robot \cite{boroson2020uwbmultiposegraph} systems. However, these methods use visual-inertial odometry (VIO), which is a simpler case (4-DoF) than ours, and only aim to improve the trajectory in offline mode. Moreover, they target different application scenarios (first exploration of an unknown environment with a single or no anchor, then offline correction after the mission is over and all data has been collected). As such, our solution is not only more complete, and can provide real-time improvement to the odometry but also can be more useful for other applications where multiple anchors can be deployed and calibrated before the mission.

In our previous works, we have conducted theoretical analyses on other range-based estimation problems such as finding the scale factor of VO and 3D position of the UWB anchor in $\VOFrame$ \cite{Thien2022scaleSDP} or finding the 4-DoF (relative 3D position and relative heading) between pairs of robots \cite{Thien2022rte4dof}. In this work, we study a different, more complex problem (7-DoF GAT, with any number of anchors), and hence the technical details are entirely new. Furthermore, the singular analyses in the aforementioned works focus only on isolated cases for each parameter and do not consider 3D rotation, while we now address the general cases for all parameters, the individual cases for each parameter, as well as including the 3D rotation.

\section{Problem Formulation} \label{sec:ProbFormulation}

\subsection{Notations} \label{subsec:notations}

Let the position vector, rotation matrix and homogeneous transformation matrix in a frame $\{\mathcal{A}\}$ be 
${\prescript{\mathcal{A}}{}{\mathbf{p}} \in \mathbb{R}^3}$, 
${\prescript{\mathcal{A}}{}{\mathbf{R}} \in SO(3)}$ and 
${\prescript{\mathcal{A}}{}{\mathbf{T}}  \coloneqq 
\begin{bmatrix} 
    \prescript{\mathcal{A}}{}{\mathbf{R}} & \prescript{\mathcal{A}}{}{\mathbf{p}} \\ 
    \mathbf{0}^\top & 1
\end{bmatrix} \in SE(3)}$, respectively. 
$\prescript{\mathcal{A}}{\mathcal{B}}{\mathbf{T}}$ and $\prescript{\mathcal{A}}{\mathcal{B}}{\mathbf{R}}$ are the transformation and rotation matrices from frame $\{\mathcal{B}\}$ to $\{\mathcal{A}\}$. 
The elements of a position vector ${\mathbf{a}} \in \R^3$ are ${\mathbf{a}} {\coloneqq} [a_x, a_y, a_z]^\top$.
$\sksym{\mathbf{a}}$ is the skew-symmetric matrix of vector $\mathbf{a}$, which is defined as
\begin{equation}
\sksym{\mathbf{a}} =
    \begin{bmatrix} 
    0    & -a_z & a_y \\
    a_z  & 0    & -a_x \\
    -a_y & a_x  & 0
    \end{bmatrix}.
\end{equation}
Let the expectation of a matrix, the noisy measurement and estimated values be $\expectation{\cdot}$, $(\tilde{\cdot})$ and $(\est{\cdot})$, respectively. The $i$-th element of a vector $\mathbf{x}$ is denoted as $x_i$, while $\mathbf{X}_{i,j}$ denotes the $(i,j)$-th element of the matrix $\mathbf{X}$.

\subsection{System Overview} \label{subsec:sys_overview}

Fig. \ref{fig:overview_operation} shows an overview of the system and the main coordinate frames.
A mobile robot, equipped with a monocular camera and a UWB sensor, traverses in an environment where $N$ UWB anchors are located. Let $\WorldFrame$, $\CameraFrame$, $\VOFrame$ and $\UWBAnchorn$ denote the world frame, camera frame, VO reference frame and $n$-th UWB anchor ($n \in [1,\cdots,N]$), respectively. $\VOFrame$ coincides with the first camera frame at initialization, i.e. ${\VOFrame \coloneqq \CameraFrame |_{t=0}}$. 
We assume that the UWB anchors' positions are pre-calibrated in $\WorldFrame$ and the anchor configuration is not singular ($N \geq 3$ not placed on the same straight line).
Let $\prescript{\mathcal{W}}{\mathcal{C}}{\mathbf{p}}$, $\prescript{\mathcal{V}}{\mathcal{C}}{\mathbf{p}}$, $\prescript{\mathcal{W}}{\UWBAnchorn}{\mathbf{p}}$ and $\prescript{\mathcal{C}}{\UWBAnchorn}{\mathbf{p}} \coloneqq 
\prescript{\mathcal{W}}{\mathcal{C}}{\mathbf{p}} - 
\prescript{\mathcal{W}}{\UWBAnchorn}{\mathbf{p}}$ be the vectors of the robot's true position, VO's up-to-scale position estimate, $\UWBAnchorn$'s true position and the true relative position between the camera and $\UWBAnchorn$ in the world frame, respectively.
Let $\RangeKNtrue$ be the true range measurement from the robot to $\UWBAnchorn$ at time instance $k$. To simplify the notations, in the rest of the paper we denote
$\prescript{\mathcal{V}}{\mathcal{C}}{\mathbf{p}}_\NumData \coloneqq \VCp_\NumData$,
$\prescript{\mathcal{W}}{\UWBAnchorn}{\mathbf{p}} \coloneqq \Wap$,
$\prescript{\mathcal{C}}{\UWBAnchorn}{\mathbf{p}} \coloneqq \Cap$.

Overall, VR-SLAM operates in three phases:
\begin{enumerate}
    \item \label{op:phaseGAT} GAT phase: the system runs VO then loosely fuse the odometry estimates with UWB ranging data to find the affine transformation matrix $\ParamGAT$
    \begin{equation}
        \ParamGAT \coloneqq
        \begin{bmatrix}
            \ParamScale \ParamRotMat & \ParamTrans \\
            \mathbf{0}^\top & 1
        \end{bmatrix}
    \end{equation}
    which comprises of a scale factor $\ParamScale$, a rotation matrix $\ParamRotMat$, and a translation vector $\ParamTrans$. 
    The relationship between the parameters and the noisy ranging measurements is
    \begin{equation}
    \begin{aligned}
        \RangeKNnoisy 
        = \RangeKNtrue + \RangeKNoise 
        = \norm{\ParamTrans + \ParamScale \AlignedPos_\NumData - \Wap} + \RangeKNoise,
    \end{aligned}
    \end{equation}
    where $\RangeKNtrue$ is assumed to be corrupted by an i.i.d. Gaussian noise $\RangeKNoise \sim \mathcal{N}(0, \sigma_{\textrm{r}}^2)$. The collected data include up-to-scale odometry data and ranging measurements:
    \begin{equation}
        \SlidingWindow = \{(
            \VCp_i^\top,
            \tilde{d}_1,\cdots
            \tilde{d}_\NumData
        )\}_{i=1,\cdots,k},
    \end{equation}
    where $\tilde{d}_i = 0$ if range data to anchor $\UWBAnchorn$ is not available (i.e., there is no LoS to $\UWBAnchorn$). Once $\tilde{\ParamGAT}$ is deemed accurate enough, the camera poses, keyframes and map points estimated in $\VOFrame$ will then be transformed into $\WorldFrame$.
    \item \label{op:phaseVRO} VRO phase: Subsequently, we estimate the camera pose and the positions of visual landmarks in $\WorldFrame$ observed in a local sliding window. By fusing both visual and ranging data in a single optimization module, VR-SLAM is able to reduce long-term drift without relying on loop closure. If there is no UWB data, VR-SLAM will work in vision-only mode but the estimates will still be in $\WorldFrame$.
    During this phase, the UPGO might be triggered to correct the drift if the UWB signal was lost intermittently.
    \item \label{op:phaseUVR} UVR phase: While VRO is running, if there is a loop closure candidate detected, the relocalization result is accepted only if the predicted ranges (i.e., the distance between the camera and UWB anchor calculated from their estimated positions) agree with the actual UWB ranges. If the tracking frontend fails, we first re-initialize the VO frontend and attempt to relocalize using images while collecting data to solve the GAT problem. If the vision-based option is not possible, the GAT problem will be solved with the last camera pose as the initial guess. If that also fails, we revert the system back to phase \ref{op:phaseGAT}.
\end{enumerate}

\section{Theoretical Analysis} \label{sec:TheorecticalAnalysis}
 
In this section, we present the derivation of the FIM ($\FIM$), the CRLB and the determinant of the FIM associated with the 7-DoF GAT problem formulated in Sect. \ref{subsec:sys_overview}. The $\FIM$ measures the amount of information encoded in a set of measurements. Hence, $\FIM$ can be a useful tool to analyze the difficulty, observability as well as to design optimal trajectory for an estimation problem \cite{ponda2009trajopt}.

\subsection{Derivation of FIM and CRLB} \label{subsec:CRLB_derivation}

A 3D rotation matrix $\ParamRotMat$ can be parameterized by a unit vector $\ParamRotAxis$, which represents the axis of rotation, and a rotation angle $\ParamRotAngle$, which describes the magnitude of the rotation about the axis, using the formula:
\begin{equation}\label{eq:angle_axis_rep}
    \ParamRotMat
    = \cos{\ParamRotAngle} \; \IdMat_{3\times3} + \sin{\ParamRotAngle} \sksym{\ParamRotAxis} + (1 - \cos{\ParamRotAngle}) \ParamRotAxis \ParamRotAxis^\top.
\end{equation}
The rotation vector $\ParamRotVec$ is defined as a vector codirectional with $\ParamRotAxis$ whose length is the rotation angle $\ParamRotAngle$, i.e. $\ParamRotVec = \ParamRotAngle \ParamRotAxis$.
The state vector, vectors of true and observed distance measurements can be written as
\begin{equation}\label{eq:state_vector}
\begin{aligned}
    &\StateVector = 
        [\ParamTrans^\top, \ParamRotVec^\top, \ParamScale]^\top
        = [t_x, t_y, t_z, v_x, v_y, v_z, \ParamScale]^\top, \\
    &\mathbf{f}(\StateVector) = 
        [d_1,\cdots,d_\NumData]^{\top}, \quad
    \tilde{\mathbf{d}} = 
        [\tilde{d}_1,\cdots,\tilde{d}_\NumData]^{\top}, \\
\end{aligned}
\end{equation}
respectively.
The $(i,j)$-th element of $\FIM$ can be written as:
\begin{equation}
    \FIM_{i,j} \coloneqq 
    \expectation{
        \frac{\partial}{\partial \Theta_i}\ln 
        \left( p(\tilde{\mathbf{d}}, \StateVector) \right)
        \frac{\partial}{\partial \Theta_j}\ln 
        \left( p(\tilde{\mathbf{d}}, \StateVector) \right)},
\end{equation}
where the natural logarithm of $p(\tilde{\mathbf{d}}, \StateVector)$ is 
\begin{equation}
    \ln \left( p(\tilde{\mathbf{d}}, \StateVector) \right) 
    = - \frac{1}{2} 
    \left(\tilde{\mathbf{d}} - \mathbf{f}(\StateVector)\right)^\top 
    \CovDR_r^{-1} 
    \left(\tilde{\mathbf{d}} - \mathbf{f}(\StateVector)\right) + c,
\end{equation}
with ${\CovDR_r = \VarR^2 \IdMat_{k \times k}}$ and $c$ being a constant. 
Assuming i.i.d. zero-mean Gaussian noise, $\FIM$ can be simplified to \cite{zekavat2011handbook}:
\begin{equation} \label{eq:FIM_original}
\begin{aligned}
    \FIM = 
    \left[ \frac{\partial \mathbf{f}(\StateVector)}{\partial \StateVector} \right]^{\top} 
    \CovDR_r^{-1} 
    \left[ \frac{\partial \mathbf{f}(\StateVector)}{\partial \StateVector} \right],
\end{aligned}
\end{equation}
where the Jacobian is
\begin{equation}
\begin{gathered}
    \frac{\partial \mathbf{f}(\StateVector)}{\partial \StateVector}
    = 
    \begin{bmatrix}
        \partial f_1 / \partial \ParamTrans & 
        \partial f_1 / \partial \ParamRotVec &
        \partial f_1 / \partial s \\
        \vdots & \vdots & \vdots \\
        \partial f_\NumData / \partial \ParamTrans & 
        \partial f_\NumData / \partial \ParamRotVec &
        \partial f_\NumData / \partial s \\
    \end{bmatrix}.
\end{gathered}
\end{equation}

The detailed computing steps for the derivatives can be found in Appendix \ref{appendix:derivatives}. Overall, we have
\begin{equation} \label{eq:derListFi}
\begin{gathered}
    \partial f_i / \partial \ParamTrans = \UnitVec_i^\top, \quad
    \partial f_i / \partial \ParamRotVec = \PartialFv_i^\top, \quad
    \partial f_i / \partial \ParamScale = \PartialFs_i \\
\end{gathered}
\end{equation}
where $i \in [1,\cdots,k]$ and
\begin{equation} \label{eq:derivativeRoi}
\begin{aligned}
    &\UnitVec_i 
    = 
    \frac{\Cap_i}{\norm{\Cap_i}}, \quad
    \PartialFv_i^\top 
    = \UnitVec_i^\top \PartialRv, \quad
    \PartialFs_i 
    = \UnitVec_i^\top \ParamRotMat \VCp_i, \\
    &\PartialRv
    =
    - \ParamScale \ParamRotMat \sksym{\VCp_i} 
    \frac{\ParamRotVec \ParamRotVec^\top + (\ParamRotMat^\top - \IdMat_{3\times3}) \sksym{\ParamRotVec}}{\norm{\ParamRotVec}^2}.
\end{aligned}
\end{equation}
In essence, $\UnitVec_i$ is a unit vector parallel to $\Cap_i$, while $\PartialRv = \partial(\ParamScale \ParamRotMat \VCp_i) / \partial \ParamRotVec$ is the derivative of $\ParamScale \ParamRotMat \VCp_i$ with respect to the exponential coordinates $\ParamRotVec$.
Eq. (\ref{eq:FIM_original}) can now be rewritten as
\begin{equation}\label{eq:FIM_simplified}
\begin{aligned}
    \FIM 
    = \frac{1}{\VarR^2} \JacobianMat^\top \JacobianMat 
    = \frac{1}{\VarR^2}
    \sum\limits_{i=1}^k
    \begin{bmatrix}
        (\frac{\partial f_i}{\partial t_x} )^2 & \dots & \frac{\partial f_i}{\partial t_x} \frac{\partial f_i}{\partial s} \\
        \vdots & \ddots & \vdots \\
        \frac{\partial f_i}{\partial s} \frac{\partial f_i}{\partial t_x} & \dots & (\frac{\partial f_i}{\partial s})^2
    \end{bmatrix},
\end{aligned}
\end{equation}
where the $i$-th row of $\JacobianMat$ is $\JacobianMat_{i,:} = 
\left[
    \partial f_i / \partial \ParamTrans, \;
    \partial f_i / \partial \ParamRotVec, \;
    \partial f_i / \partial \ParamScale
\right]$.

The CRLB defines the lower bound of the achievable error covariance for an unbiased estimator and can be computed if $\detF \neq 0$, or $\FIM$ is non-singular:
\begin{equation} \label{ch03:eq:CRLB_definition}
    \expectation{(\est{\StateVector} - \StateVector) (\est{\StateVector} - \StateVector)^{\top}} \geq \textrm{CRLB} = \FIM^{-1}.
\end{equation}
Since the geometry of the anchors and the robot's trajectory are the main factors that affect the CRLB and not the estimation method itself \cite{ponda2009trajopt}, it can be used to measure system performance. 

\subsection{Determinant of the FIM} \label{subsec:det_FIM}

Applying the Cauchy-Binet formula on Eq. (\ref{eq:FIM_simplified}), we get
\begin{equation}\label{ch03:eq:det_FIM_Cauchy}
    \detF = \frac{1}{\VarR^2} \sum\limits_{S}^{} \left(\detL\right)^2,
\end{equation}
where ${S = \{ (m_1,\cdots,m_7) \; \vert \; 1 \leq m_1 < \cdots < m_7 \leq k\}}$
and
$\SubFIM$ is a ${7 \times 7}$ matrix consisting of $m_1, \cdots, m_7$-th rows of $\JacobianMat$.
To improve readability, we will analyze $\detL$ with the first 7 rows of $\JacobianMat$ and then generalize the results.

Denote $\VEdge = [\UnitVec^\top, \PartialFv^\top, \PartialFs]^\top \in \R^7$, we have:
\begin{equation} \label{eq:Lambda_17}
    \SubFIM \coloneqq 
    \begin{bmatrix}
        \UnitVec_1^\top & \PartialFv_1^\top & \PartialFs_1 \\
        \vdots & \vdots & \vdots \\
        \UnitVec_7^\top & \PartialFv_7^\top & \PartialFs_7
    \end{bmatrix}
    =
    \begin{bmatrix}
        \VEdge_1^\top \\
        \vdots \\
        \VEdge_7^\top
    \end{bmatrix}.
\end{equation}

Another way to analyze $\detL$ is to apply the Laplace expansion along the last column of $\SubFIM$, which leads to:
\begin{equation}\label{ch03:eq:det_Lambda_Laplace_expansion}
\begin{aligned}
    \detL
    &= \sum\limits_{i=1}^{7} (-1)^i \PartialFs_i \det(\SubFIM_{i7}), \\
\end{aligned}
\end{equation}
where $\SubFIM_{i7}$ is the submatrix obtained by removing row $i$ and the last column of $\SubFIM$. We can then express $\det(\SubFIM_{i7})$ by utilizing its $3\times3$ blocks. For example, with $i = 7$:
\begin{equation}
    \SubFIM_{77} = 
    \begin{bmatrix}
        \UnitVec_1^\top & \PartialFv_1^\top \\
        \vdots & \vdots \\
        \UnitVec_6^\top & \PartialFv_6^\top
    \end{bmatrix}
    =
    \begin{bmatrix}
        \mathbf{A}_1 & \mathbf{A}_2 \\
        \mathbf{A}_3 & \mathbf{A}_4
    \end{bmatrix}
\end{equation}
where $\mathbf{A}_1,\cdots,\mathbf{A}_4$ are the $3 \times 3$ blocks of $\SubFIM_{77}$. 
Assuming $\mathbf{A}_1$ is invertible, $\det(\SubFIM_{77})$ can be computed as
\begin{equation}
\begin{aligned}
    \det(\SubFIM_{77}) = 
    \det(\mathbf{A}_4 - \mathbf{A}_3 \mathbf{A}_1^{-1} \mathbf{A}_2) \det(\mathbf{A}_1),
\end{aligned}
\end{equation}
where 
${\det(\mathbf{A}_1) = \mdet{\UnitVec_1 \; \UnitVec_2 \; \UnitVec_3}
= (\mathbf{u}_1 \times \mathbf{u}_2) \cdot \mathbf{u}_3}$ and
\begin{equation}
\begin{aligned}
    &\det(\mathbf{A}_4 - \mathbf{A}_3 \mathbf{A}_1^{-1} \mathbf{A}_2) \\
    &= 
    \mdet{ 
    \begin{bmatrix}
        \PartialFv_4^\top \\
        \PartialFv_5^\top \\
        \PartialFv_6^\top
    \end{bmatrix}
    - \dfrac{1}{\det(\mathbf{A}_1)}
    \begin{bmatrix}
        \UnitVec_4^\top \\
        \UnitVec_5^\top \\
        \UnitVec_6^\top
    \end{bmatrix}
    \begin{bmatrix}
        \UnitVec_2^\top \times \UnitVec_3^\top \\
        \UnitVec_3^\top \times \UnitVec_1^\top \\
        \UnitVec_1^\top \times \UnitVec_2^\top
    \end{bmatrix}
    \begin{bmatrix}
        \PartialFv_1^\top \\
        \PartialFv_2^\top \\
        \PartialFv_3^\top
    \end{bmatrix}
    }.
\end{aligned}
\end{equation}

Combining the above equations, we have
\begin{align}
    &\detF 
    = \frac{1}{\VarR^2} \sum\limits_{S}^{} \left(\detL\right)^2 
    = \frac{1}{\VarR^2} \sum\limits_{S}^{} \left( \det[\lambda_{p,q}] \right)^2 \label{eq:detFMath} \\
    &= 
    \frac{1}{\VarR^2} \sum\limits_{S}^{} 
    \left[
        \sum\limits_{i=1}^{7} (-1)^i 
        \PartialFs_{m_i} \det(\SubFIM_{i7}) 
    \right]^2 \\
    &= 
    \frac{1}{\VarR^2} \sum\limits_{S}^{} 
    \left[
        \sum\limits_{i=1}^{7} (-1)^i 
        \PartialFs_{m_i}
        \DetSubRot_i 
        \DetSubTrans_i
    \right]^2 \label{eq:detFPhys}
\end{align}
where ${S = \{ (m_1,\cdots,m_7) \; \vert \; 1 \leq m_1 < \cdots < m_7 \leq k\}}$, $m_1 \leq p,q \leq m_7$,
\begin{equation}\label{eq:detABC}
\begin{aligned}
    &\lambda_{p,q} = \VEdge_p \cdot \VEdge_q = \UnitVec_p \cdot \UnitVec_q + \PartialFv_p \cdot \PartialFv_q+ \PartialFs_p \PartialFs_q, \\
    &\DetSubTrans_i =
    (\mathbf{u}_{l_1} \times \mathbf{u}_{l_2}) \cdot \mathbf{u}_{l_3},
    \quad \PartialFs_i = \UnitVec_i^\top \ParamRotMat \VCp_i, \\
    &\DetSubRot_i = 
    \mdet{ 
    \begin{bmatrix}
        \PartialFv_{l_4}^\top \\
        \PartialFv_{l_5}^\top \\
        \PartialFv_{l_6}^\top
    \end{bmatrix}
    - \dfrac{1}{\DetSubTrans_i}
    \begin{bmatrix}
        \UnitVec_{l_4}^\top \\
        \UnitVec_{l_5}^\top \\
        \UnitVec_{l_6}^\top
    \end{bmatrix}
    \begin{bmatrix}
        \UnitVec_{l_2}^\top \times \UnitVec_{l_3}^\top \\
        \UnitVec_{l_3}^\top \times \UnitVec_{l_1}^\top \\
        \UnitVec_{l_1}^\top \times \UnitVec_{l_2}^\top
    \end{bmatrix}
    \begin{bmatrix}
        \PartialFv_{l_1}^\top \\
        \PartialFv_{l_2}^\top \\
        \PartialFv_{l_3}^\top
    \end{bmatrix}
    }, \\
\end{aligned}
\end{equation}
with $\{l_1, \cdots, l_6\} \subseteq \{m_1,\cdots,m_7\} \setminus m_i$,
$\UnitVec$ and $\PartialFv$ are defined in Eq. (\ref{eq:derListFi}-\ref{eq:derivativeRoi}). Note that if $\DetSubTrans_i = 0$ then $\det(\SubFIM_{i7}) = 0$ and $\DetSubRot_i$ becomes irrelevant.

\subsection{Geometric Interpretation} \label{subsec:GeoAnalysis}

From Eq. (\ref{eq:detFMath}), it can be seen that for every set of seven measurements in $S$, $\detL$ is the signed volume of a parallelepiped in $7$ dimensions with $\VEdge_p,\cdots,\VEdge_q$ as edges. $\detF$ is the sum of the squared volume over all possible combinations of measurements. Given that $\detF$ is inversely proportional to the volume of uncertainty, smaller $\detF$ means larger achievable variances of the estimates, or the problem becomes harder. If $\detF$ is zero, then no fully $7$-dimensional parallelepiped can be generated from the available measurements or the system is unobservable.

Eq. (\ref{eq:detFPhys}) gives a clearer physical interpretation of the main components of $\detF$. 
Firstly, $\PartialFs_i$, $\DetSubRot_i$, and $\DetSubTrans_i$ represent the information gain with respect to the scale, rotation and translation parts of $\StateVector$. $\detL$ is a combination of these components over one set of 7 measurements, then $\detF$ is the sum of all $\detL^2$ over all possible sets of measurements in $S$.
Secondly, with larger $S$ (i.e., more measurements collected) or smaller $\VarR$ (i.e., more accurate measurements), $\detF$ becomes larger and the uncertainty generally becomes smaller.
Thirdly, the key factors that affect the value of $\detF$ include:
\begin{itemize}
    \item When the robot does not move ($\norm{\VCp_i} = 0 \; \forall i$) or follows a trajectory such that $\UnitVec_i$ and $\ParamRotMat \VCp_i$ are perpendicular (i.e., $\UnitVec_i^\top \ParamRotMat \VCp_i = 0$), then $\PartialFs_i = 0$. Similarly, if the anchors and robot's trajectory are on the same 2D plane (all $\UnitVec_i$ vectors are coplanar), then $\DetSubTrans_i = 0$. Both cases lead to $\detF = 0$ and the system is unobservable.
    \item When $\ParamScale$ is very large, $\norm{\VCp_i}$ will be very small and so will $\PartialFs_i$. In that case, the last column of $\detF$ will be close to zero and the system is unobservable.
    \item $\DetSubTrans_i$ depends on the spanning angles between the $\UnitVec_{l_1}$, $\UnitVec_{l_2}$, $\UnitVec_{l_3}$ vectors but not on the measurement values ($d_{l_1}$, $\cdots$, $d_{l_3}$) or sensing positions ($\WCp_{l_1}$ $\cdots$, $\WCp_{l_3}$). Hence, circling around the target should generally improve the estimates, which is a known result in the range-based relative localization literature \cite{guo2017reloc}.
\end{itemize} 

\subsection{Singular Configuration Analysis} \label{subsec:SingularAnalysis}

\begin{figure}[t]
    \begin{subfigure}[t]{.48\linewidth}
    \centering
    \includegraphics[width=\linewidth]{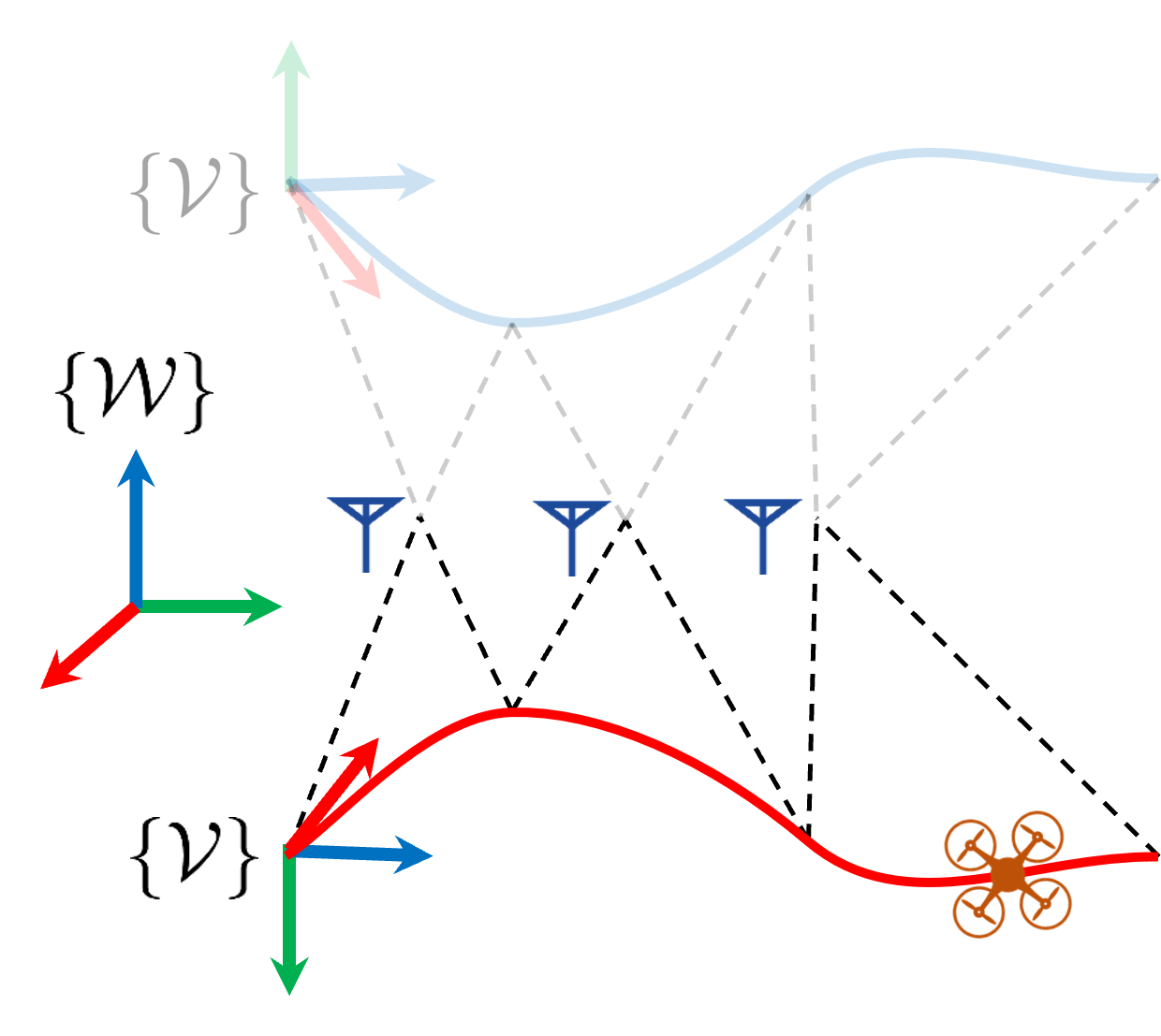}
    \caption{Unobservable translation}
    \label{fig:singular_t}
    \end{subfigure}
    \hfill
    \begin{subfigure}[t]{.48\linewidth}
    \centering
    \includegraphics[width=\linewidth]{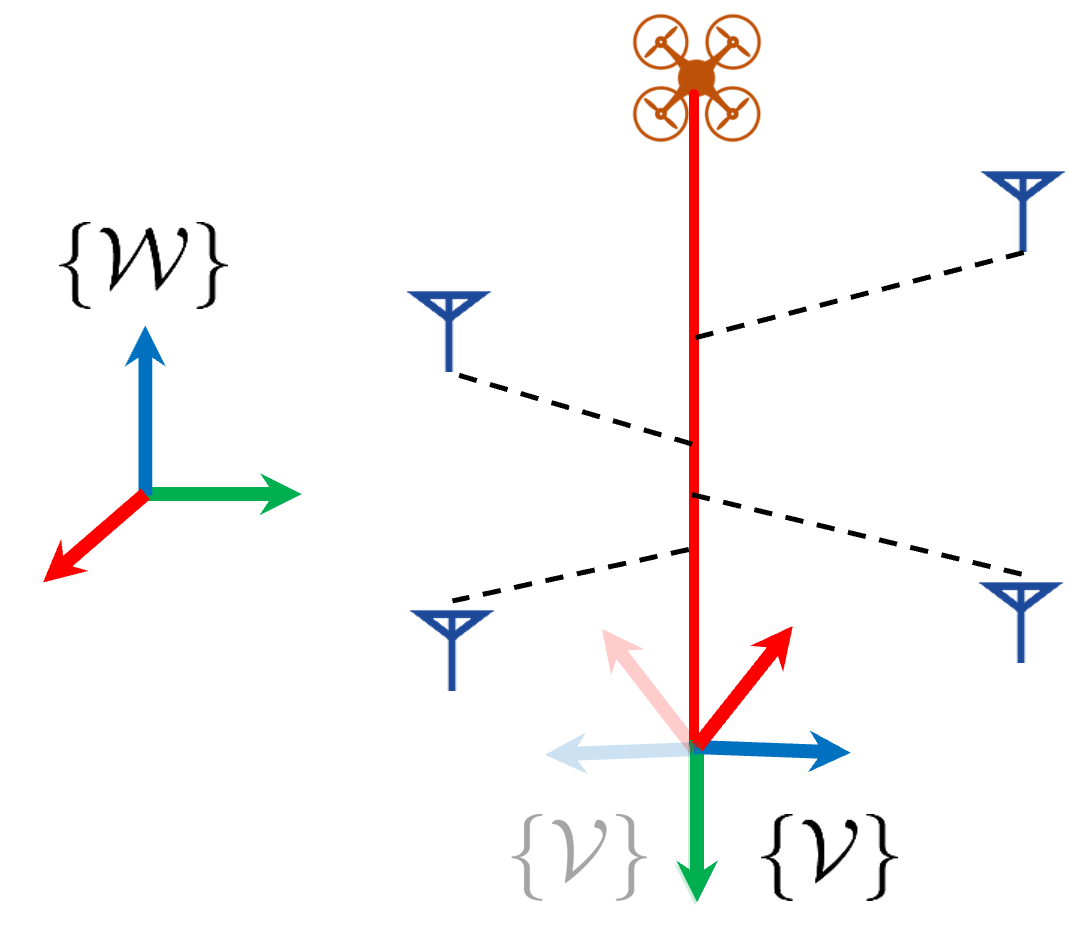}
    \caption{Unobservable rotation}
    \label{fig:singular_R}
    \end{subfigure}
    \caption{Examples of singular configurations. The true and possible solutions are transparent and solid lines, respectively.}
    \label{fig:singular_exp}
\end{figure}

The configurations that result in $\FIM$ becomes singular are also unobservable cases \cite{wang2008FIMobsSLAM}. Since $\FIM = \frac{1}{\VarR^2} \JacobianMat^\top \JacobianMat$, we have $\rank(\FIM) = \rank(\JacobianMat)$ and can study the observability of the system through $\JacobianMat$. In particular, $\FIM$ is non-singular if and only if the Jacobian $\JacobianMat \in \mathbb{R}^{\NumData \times 7}$ is full column rank, where $k$ is the number of measurements. 

If there is an insufficient number of measurements, i.e. $\NumData < 7$, then FIM cannot be full rank. If $\NumData \geq 7$, $\JacobianMat$ is singular if and only if its column vectors are linearly dependent, i.e. there exists a non-trivial solution $\SolVector \in \mathbb{R}^{7}$ to the following overdetermined homogeneous linear system:
\begin{align*}
    &\JacobianMat \SolVector = \mathbf{0}
\Leftrightarrow
    \begin{bmatrix}
        \UnitVec_1^\top & \PartialFv_1^\top & \PartialFs_1 \\
        \vdots & \vdots & \vdots \\
        \UnitVec_\NumData^\top & \PartialFv_\NumData^\top & \PartialFs_\NumData
    \end{bmatrix}
    \SolVector
    = \mathbf{0} \\
\Leftrightarrow
    &\begin{bmatrix}
        \UnitVec_1^\top 
        & -\ParamScale \UnitVec_1^\top \ParamRotMat \sksym{\VCp_1} \DerivativeMat 
        & \UnitVec_1^\top \ParamRotMat \VCp_1 \\
        \vdots & \vdots & \vdots \\
        \UnitVec_\NumData^\top 
        & -\ParamScale \UnitVec_\NumData^\top \ParamRotMat \sksym{\VCp_\NumData} \DerivativeMat
        & \UnitVec_\NumData^\top \ParamRotMat \VCp_1
    \end{bmatrix}
    \begin{bmatrix}
        \SolVector_{1:3} \\
        \SolVector_{4:6} \\
        c_7
    \end{bmatrix}
    = \mathbf{0}
\end{align*}
\begin{equation} \label{eq:SingularGeneral}
    \Leftrightarrow
    \UnitVec_i \cdot (\SolVector_{1:3} 
    - \ParamScale \ParamRotMat \sksym{\VCp_i} \DerivativeMat \SolVector_{4:6} 
    + c_7 \ParamRotMat \VCp_i) = 0 \; (1 \leq i \leq k) \\
\end{equation}
where $\SolVector_{a:b}$ is the vector of $a$-th to $b$-th elements of $\SolVector$ and
\begin{equation}
\DerivativeMat = 
    \frac{\ParamRotVec \ParamRotVec^\top + (\ParamRotMat^\top - \IdMat_{3\times3}) \sksym{\ParamRotVec}}{\norm{\ParamRotVec}^2}.
\end{equation}

In essence, the overdetermined homogeneous linear system (OHLS) (\ref{eq:SingularGeneral}) encompasses all singular cases and can be used to check for unobservability of a configuration (defined by the true transformation $\ParamGAT$, all unit relative position vectors $\UnitVec_i$ and odometry data $\VCp_i$). However, it is difficult to understand the singular conditions for the translation, rotation and scale parameters individually. Hence, we isolate each parameter and analyze the corresponding degenerate cases in the rest of this section. An arbitrary unobservable configuration would be a combination of these cases. Fig. \ref{fig:singular_exp} illustrates some of these singular cases. 

\subsubsection{Unobservable translation}
By setting $\SolVector_{4:7} = \mathbf{0}$, the OHLS (\ref{eq:SingularGeneral}) can be rewritten as
\begin{equation}
    \UnitVec_i \cdot \SolVector_{1:3} = 0  \quad \forall i \in [1, \cdots, \NumData]
\end{equation}
which has a non-trivial solution $\SolVector_{1:3}$ if and only if the set $\{\UnitVec_i\}_{i=1,\cdots,\NumData}$ is a subspace of $\R^3$. Intuitively, this means the vectors $\UnitVec_i$ are all parallel or on the same 2D plane. Hence, the necessary and sufficient condition for $\ParamTrans$ to be observable is that all $\UnitVec_i$ vectors are not coplanar. This result also applies if the last 4 columns of $\JacobianMat$ are removed, i.e. $\ParamRotMat$ and $\ParamScale$ are known.

\subsubsection{Unobservable rotation}
By setting $\SolVector_{1:3} = \mathbf{0}$ and $c_7 = 0$, the OHLS (\ref{eq:SingularGeneral}) can be rewritten as
\begin{equation} \label{eq:SingularRot01}
    \UnitVec_i \cdot (\ParamScale \ParamRotMat \sksym{\VCp_i} \DerivativeMat \SolVector_{4:6}) = 0  \quad \forall i \in [1, \cdots, \NumData].
\end{equation}
Given a vector $\mathbf{a} \in \R^3$ and a $3\times3$ invertible matrix $\mathbf{G}$, we have a cross product relation:
\begin{equation}
    \sksym{\mathbf{G} \mathbf{a}} = \det(\mathbf{G}) \mathbf{G}^{-\top} \sksym{\mathbf{a}} \mathbf{G}^{-1}.
\end{equation}
Using this relation, we can write:
\begin{equation}
\begin{aligned}
    &\sksym{\ParamRotMat \VCp_i} 
    = \det(\ParamRotMat) \ParamRotMat^{-\top} \sksym{\VCp_i} \ParamRotMat^{-1}
    = \ParamRotMat \sksym{\VCp_i} \ParamRotMat^\top \\
    &\Rightarrow \label{eq:SimplifiedsRoi}
    \sksym{\ParamScale \ParamRotMat \VCp_i} \ParamRotMat
    = \ParamScale \ParamRotMat  \sksym{\VCp_i}
\end{aligned}
\end{equation}
Denote $\BodyVec_i \coloneqq \ParamScale \ParamRotMat \VCp_i$, which is essentially the true translation vector of the robot from its starting point, and combine with Eq. (\ref{eq:SingularRot01})-(\ref{eq:SimplifiedsRoi}), we have:
\begin{equation} \label{eq:SingularRot02}
    \UnitVec_i \cdot (\sksym{\BodyVec_i} \ParamRotMat \DerivativeMat \SolVector_{4:6}) = 0  \quad \forall i \in [1, \cdots, \NumData].
\end{equation}
Since $\ParamRotMat$ and $\DerivativeMat$ are constant matrices, we can effectively combine them with $\SolVector_{4:6}$ and rewrite (\ref{eq:SingularRot02}) as 
\begin{equation} \label{eq:SingularRot03}
\begin{aligned}
    &\UnitVec_i \cdot (\sksym{\BodyVec_i} \SolVector') = 0  \quad 
    \Leftrightarrow
    \UnitVec_i \cdot (\BodyVec_i \cross \SolVector') = 0 \\
    &\Leftrightarrow
    (\UnitVec_i \cross \BodyVec_i) \cdot \SolVector'= 0 \quad \forall i \in [1, \cdots, \NumData] 
\end{aligned}
\end{equation}
where $\SolVector' = \ParamRotMat \DerivativeMat \SolVector_{4:6}$. The OHLS (\ref{eq:SingularRot03}) has a non-trivial solution $\SolVector'$ if and only if the set $\{\UnitVec_i \cross \BodyVec_i\}_{i=1,\cdots,\NumData}$ is a subspace of $\R^3$.
Intuitively, this means that we can find a non-zero vector $\SolVector'$ that is perpendicular to all $\{\UnitVec_i \cross \BodyVec_i\}$ vectors. Hence, the necessary and sufficient condition for $\ParamTrans$ to be observable is that all $\{\UnitVec_i \cross \BodyVec_i\}$ vectors are not coplanar.

\subsubsection{Unobservable scale}
By setting $\SolVector_{1:6} = \mathbf{0}$, the OHLS (\ref{eq:SingularGeneral}) can be rewritten as (note that $c_7 \neq 0$, $d_i > 0$ and $\ParamScale > 0$):
\begin{equation} \label{eq:SingularScale01}
\begin{aligned}
    &\UnitVec_i \cdot (c_7 \ParamRotMat \VCp_i) = 0 
    \Leftrightarrow
    (d_i \UnitVec_i) \cdot (\ParamScale \ParamRotMat \VCp_i) = 0 \\
    &\Leftrightarrow
    \Cap_i \cdot \BodyVec_i = 0 \quad \forall i \in [1, \cdots, \NumData].
\end{aligned}
\end{equation}
Essentially, we can choose any $c_7 \neq 0$ as the solution so long as the robot trajectory and the anchor positions are such that $\Cap_i$ and $\BodyVec_i$ are always perpendicular, then $\ParamScale$ is unobservable. We can replace $\BodyVec_i = \WCp_i - \ParamTrans$ and $\Cap_i = \Wap - \WCp_i$ in each equation of (\ref{eq:SingularScale01}) to get:
\begin{equation} \label{eq:SingularScale02}
\begin{aligned}
    &(\Wap - \WCp_i) \cdot (\WCp_i - \ParamTrans) = 0 \\
    \Leftrightarrow
    &(\WCp_i - \ParamTrans)^\top (\WCp_i - \ParamTrans + \ParamTrans - \Wap) = 0 \\
    \Leftrightarrow
    &\norm{\WCp_i - \ParamTrans}^2 + (\ParamTrans - \Wap) \cdot (\WCp_i - \ParamTrans) \\
    &+ \frac{1}{4} \norm{\ParamTrans - \Wap}^2 - \frac{1}{4} \norm{\ParamTrans - \Wap}^2 = 0 \\
    \Leftrightarrow
    &\norm{\WCp_i - \ParamTrans + \frac{1}{2}(\ParamTrans - \Wap)}^2 = \frac{1}{4} \norm{\ParamTrans - \Wap}^2  \\
    \Leftrightarrow
    &\norm{\WCp_i - \frac{1}{2}(\ParamTrans + \Wap)} = \frac{1}{2} \norm{\ParamTrans - \Wap}  \\
\end{aligned}
\end{equation}
Intuitively, Eq. (\ref{eq:SingularScale02}) means that at time instance $i$, the robot measures range data to $\UWBAnchorn$ while moving on the surface of a sphere centered at $\frac{1}{2} (\ParamTrans + \Wap)$ with radius $\frac{1}{2} \norm{\ParamTrans - \Wap}$, i.e. the sphere with the line connecting the starting point and anchor $\UWBAnchorn$ as the diameter.

\section{Proposed System} \label{sec:ProposedSystem}

\begin{figure}[t]
\centering
\includegraphics[width=\linewidth]{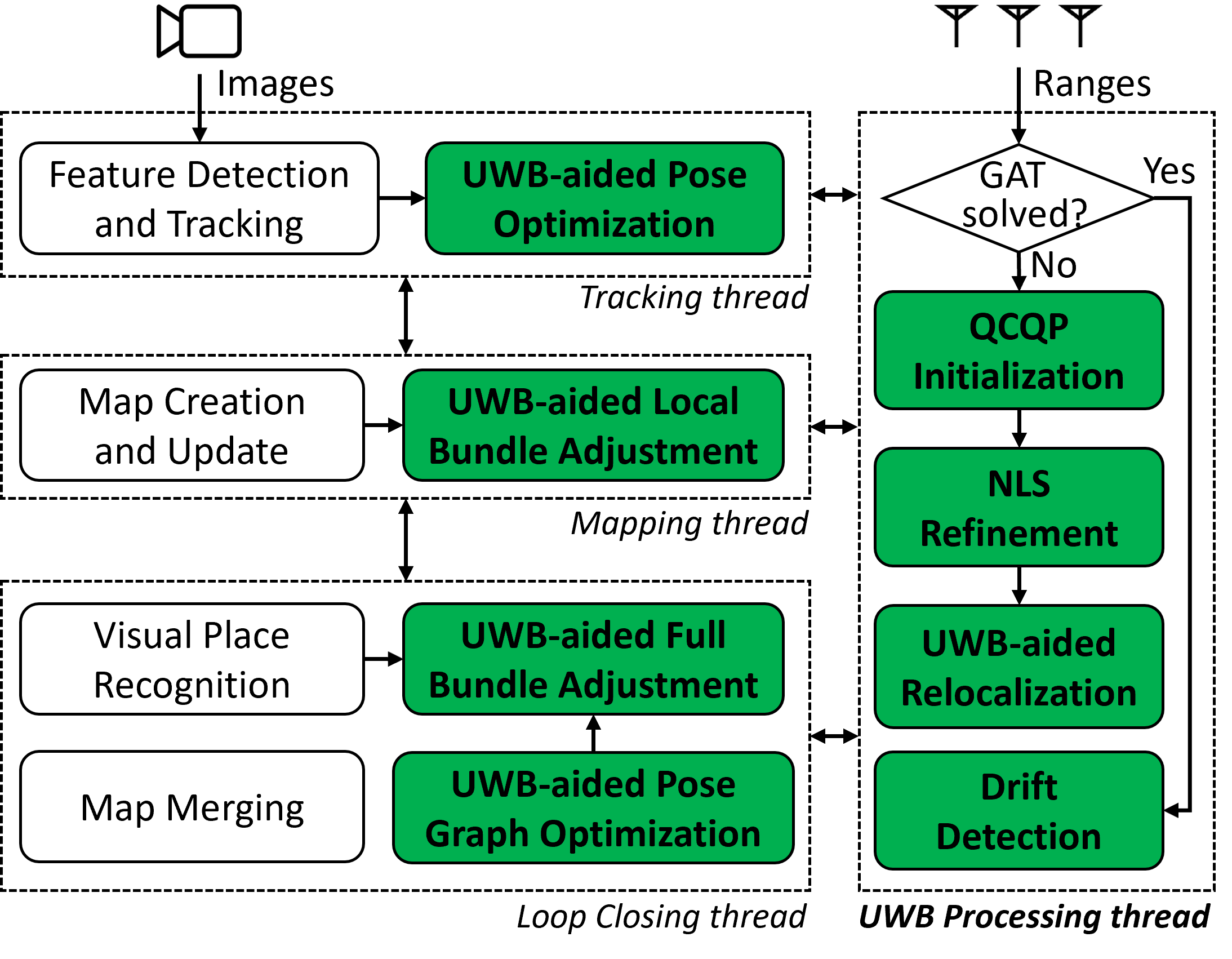}
    \caption{System architecture of VR-SLAM. The main components proposed in this work are highlighted.}
    \label{fig:vrslam_pipeline}
\end{figure}

In this section, we present the new components of the proposed VR-SLAM system. An overview of VR-SLAM's processing pipeline is illustrated in Fig. \ref{fig:vrslam_pipeline}. 
We refer to ORB-SLAM2 \cite{mur2017orb} for the details regarding the vision-only modules.

\subsection{Global Affine Transformation (GAT) Estimation} 

\subsubsection{Initialization with QCQP} \label{subsec:ProposedSystem_SD_WLS}

The true distance and the square measurement can be written as
\begin{equation}
\begin{aligned}
    d_\NumData 
    = \norm{\Cap_\NumData} = \sqrt{\Cap_\NumData^\top \Cap_\NumData}, \;\;
    \tilde{d}^2_\NumData 
    = d_\NumData^2 + 2 d_\NumData \RangeKNoise + \RangeKNoise^2.
\end{aligned}
\end{equation}
Let $\nu_\NumData \coloneqq 2 d_\NumData \RangeKNoise + \RangeKNoise^2$ be the noise term for $\tilde{d}^2_\NumData$. Although $\nu_\NumData$ is not zero-mean Gaussian, it can be approximated as one \cite{trawny2010relplanar}:
\begin{equation}
\begin{aligned}
    &\delta_\NumData = d_\NumData^2 + \eta'_\NumData = \Cap_\NumData^\top \Cap_\NumData + \eta'_\NumData,\\
    &\delta_\NumData \simeq \tilde{d}_\NumData^2 - \bar{\nu}_\NumData, \;\; \bar{\nu}_\NumData \coloneqq \expectation{\nu_\NumData} = \CovMatR_{\NumData,\NumData},\\
    &\bm{\eta}' = [\eta'_1, \dots, \eta'_\NumData] \sim \mathcal{N}(\mathbf{0}, \CovMatS).
\end{aligned}
\end{equation}

Firstly, $\ParamGAT$ can be estimated by minimizing the square distances-weighted least squares cost function  \cite{trawny2010relplanar}:
\begin{equation}\label{eq:prob_SD_WLS}
    \min\limits_{\ParamGAT} \frac{1}{2}
        \mathbf{e}_s^\top
        \CovMatS^{-1}
        \mathbf{e}_s
\end{equation}
where the vector of the squared distance errors is
\begin{equation} \label{eq:esquare}
    \mathbf{e}_s \coloneqq [
    \Cap_1^\top \Cap_1 - \delta_1, \;
    \dots, \;
    \Cap_\NumData^\top \Cap_\NumData - \delta_\NumData]^\top.
\end{equation}
By expanding and simplifying $\Cap_k^\top \Cap_k$, we have:
\begin{align*}
    &\Cap_k^\top \Cap_k 
    = (\ParamTrans + \ParamScale \AlignedPos_\NumData - \Wap)^\top (\ParamTrans + \ParamScale \AlignedPos_\NumData - \Wap) \\
    &=
    \norm{\ParamTrans}^2 + \ParamScale^2 \norm{\VCp_k}^2 + \norm{\Wap}^2 + 
    2 \ParamScale (\ParamTrans - \Wap)^\top \AlignedPos_\NumData - 
    2 \Wap^\top \ParamTrans \\
    &= \QCQPLeftMat_k \QCQPStateVec,
\end{align*}
where
\begin{align*}
    &\QCQPStateVec \coloneqq 
    [
    \ParamTrans^\top,
    \norm{\ParamTrans}^2,
    \ParamScale^2,
    \ParamScale \ParamRasVec^\top,
    \ParamScale \ParamTrans^\top \ParamRotMat, 
    1
    ]^\top \in \R^{18 \times 1}, \stepcounter{equation}\tag{\theequation}\label{eq:x_state_vector}\\
    &\QCQPLeftMat_k \coloneqq 
    [
    - 2 \Wap^\top, \quad
    1, \quad
    \norm{\VCp_k}^2, \quad
    - 2 (\Wap^\top \UnitX) \VCp_k^\top, \\
    &- 2 (\Wap^\top \UnitY) \VCp_k^\top, \;
    - 2 (\Wap^\top \UnitZ) \VCp_k^\top, \;
    2 \VCp_k^\top, \;
    \norm{\Wap}^2 - \delta_k
    ] \in \R^{1 \times 18}.
\end{align*}
Here, $\ParamRasVec = [\ParamRotMat_{1,1}, \ParamRotMat_{1,2}, \cdots, \ParamRotMat_{9,9}]^\top$ is the vector which contains all the elements of $\ParamRotMat$, $\UnitVec_{x,y,z}$ are the unit vectors along the $x$,$y$,$z$ axes in $\R^3$. In essence, we have separated the data in $\QCQPLeftMat_k$ and the unknown parameters in $\QCQPStateVec$.
The cost function (\ref{eq:prob_SD_WLS}) can now be rewritten as
\begin{equation}
    \frac{1}{2} \mathbf{e}_s^\top \CovMatS^{-1} \mathbf{e}_s 
    = 
    \frac{1}{2} \QCQPStateVec^\top 
    \overbrace{
    \begin{bmatrix}
        \QCQPLeftMat_1\\
        \vdots \\
        \QCQPLeftMat_k
    \end{bmatrix}^\top
    \CovMatS^{-1} 
    \begin{bmatrix}
        \QCQPLeftMat_1\\
        \vdots \\
        \QCQPLeftMat_k
    \end{bmatrix}
    }^{\QCQPConsMatrix_0}
    \QCQPStateVec 
    = \frac{1}{2} \QCQPStateVec^\top \QCQPConsMatrix_0 \QCQPStateVec.
\end{equation}

Finally, we reformulate the problem (\ref{eq:prob_SD_WLS}) as a non-convex QCQP problem
\begin{equation}\label{eq:prob_QCQP}
\begin{aligned}
    \min\limits_{\QCQPStateVec} \;\; & \QCQPStateVec^\top \QCQPConsMatrix_0 \QCQPStateVec\\
    \textrm{s.t.} \;\; 
    &\QCQPStateVec^\top \QCQPConsMatrix_i \QCQPStateVec = \QCQPConsScalar_i,\;\;i=1,\dots,13,\\
\end{aligned}
\end{equation}
where the constraints are obtained from the relationship between the parameters as well as the orthogonality of the rotation matrix $\ParamRotMat$.
We use the MATLAB operator  $\sparse{\mathbf{a},\mathbf{b},\mathbf{c},m,n}$ which generates an $m \times n$ sparse matrix $\QCQPConsMatrix$ from the vectors $\mathbf{a}$, $\mathbf{b}$, and $\mathbf{c}$, such that $\QCQPConsMatrix_{a_k,b_k} = c_k$.

The first 7 constraints for the QCQP problem (\ref{eq:prob_QCQP}) can be obtained from the relations between the parameters:
\begin{itemize}
    \item $t_x^2 + t_y^2 + t_z^2 = \norm{\ParamTrans}^2$ \\
        $\Leftrightarrow
        x_1^2 + x_2^2 + x_3^2 = x_4 x_{18}$ 
        $\Rightarrow 
        \QCQPStateVec^\top \QCQPConsMatrix_1 \QCQPStateVec = \QCQPConsScalar_1$,
    \item ${(\ParamScale \ParamTrans^\top \ParamRotMat) (\ParamScale \ParamTrans^\top \ParamRotMat)^\top 
    = \ParamScale^2 \ParamTrans^\top \ParamRotMat \ParamRotMat^\top \ParamTrans
    = \ParamScale^2 \norm{\ParamTrans}^2}$
    ${\Leftrightarrow
    x_{15}^2 + x_{16}^2 + x_{17}^2 - x_4 x_5 = 0}$
    $\Rightarrow 
    \QCQPStateVec^\top \QCQPConsMatrix_2 \QCQPStateVec = \QCQPConsScalar_2$,
    \item ${\ParamScale (r_1 t_x + r_4 t_y + r_7 t_z) =
    \ParamScale \ParamTrans^\top \ParamRotMat \UnitX}$ 
    ${\Leftrightarrow
    x_1 x_6 + x_2 x_9 + x_3 x_{12} = x_{15} x_{18}}$
    $\Rightarrow 
    \QCQPStateVec^\top \QCQPConsMatrix_3 \QCQPStateVec = \QCQPConsScalar_3$,
    \item ${\ParamScale (r_2 t_x + r_5 t_y + r_8 t_z) =
    \ParamScale \ParamTrans^\top \ParamRotMat \UnitY}$ 
    ${\Leftrightarrow
    x_1 x_7 + x_2 x_{10} + x_3 x_{13} = x_{16} x_{18}}$
    $\Rightarrow 
    \QCQPStateVec^\top \QCQPConsMatrix_4 \QCQPStateVec = \QCQPConsScalar_4$,
    \item ${\ParamScale (r_3 t_x + r_6 t_y + r_9 t_z) =
    \ParamScale \ParamTrans^\top \ParamRotMat \UnitZ}$ 
    ${\Leftrightarrow
    x_1 x_8 + x_2 x_{11} + x_3 x_{14} = x_{17} x_{18}}$
    $\Rightarrow 
    \QCQPStateVec^\top \QCQPConsMatrix_5 \QCQPStateVec = \QCQPConsScalar_5$,
    \item $\norm{\ParamTrans}^2 = d_0^2$ $\Leftrightarrow$ $x_4 x_{18} = d_0^2$
    $\Rightarrow 
    \QCQPStateVec^\top \QCQPConsMatrix_6 \QCQPStateVec = \QCQPConsScalar_6$,
    \item $x_{18} = 1$
    $\Rightarrow 
    \QCQPStateVec^\top \QCQPConsMatrix_7 \QCQPStateVec = \QCQPConsScalar_7$,
\end{itemize}
where
\begin{align*} 
    &\QCQPConsMatrix_1 = \sparse{[1,2,3,4],[1,2,3,18],[1,1,1,-1],18,18},\\
    &\QCQPConsMatrix_2 = \sparse{[15,16,17,4],[15,16,17,5],[1,1,1,-1],18,18},\\
    &\QCQPConsMatrix_3 = \sparse{[1,2,3,15],[6,9,12,18],[1,1,1,-1],18,18},\\
    &\QCQPConsMatrix_4 = \sparse{[1,2,3,16],[7,10,13,18],[1,1,1,-1],18,18},\\
    &\QCQPConsMatrix_5 = \sparse{[1,2,3,17],[8,11,14,18],[1,1,1,-1],18,18},\\
    &\QCQPConsMatrix_6 = \sparse{[4],[18],[1],18,18},\tag{\stepcounter{equation}\theequation}\\
    &\QCQPConsMatrix_7 = \sparse{[18],[18],[1],18,18},\\
    &\QCQPConsScalar_1 = 0, \;
    \QCQPConsScalar_2 = 0, \;
    \QCQPConsScalar_3 = 0, \;
    \QCQPConsScalar_4 = 0, \;
    \QCQPConsScalar_5 = 0,\;
    \QCQPConsScalar_6 = \tilde{d}_0^2,\;
    \QCQPConsScalar_7 = 1. 
\end{align*}
Here, $\tilde{d}_0$ is the distance between the origin of $\WorldFrame$ to the robot's starting position, which can be measured during the calibration process.
The next 6 constraints come from the orthogonality of the rotation matrix $\ParamRotMat$:
\begin{itemize}
    \item $\ParamRotMat_{1,1}^2 + \ParamRotMat_{2,1}^2 + \ParamRotMat_{3,1}^2 = 1$ 
    $\Leftrightarrow x_6^2 + x_9^2 + x_{12}^2 = x_5 x_{18}$\\
    $\Rightarrow 
    \QCQPStateVec^\top \QCQPConsMatrix_8 \QCQPStateVec = \QCQPConsScalar_8$,
    \item $\ParamRotMat_{1,2}^2 + \ParamRotMat_{2,2}^2 + \ParamRotMat_{3,2}^2 = 1$ 
    $\Leftrightarrow x_7^2 + x_{10}^2 + x_{13}^2 = x_5 x_{18}$\\
    $\Rightarrow 
    \QCQPStateVec^\top \QCQPConsMatrix_9 \QCQPStateVec = \QCQPConsScalar_9$,
    \item $\ParamRotMat_{1,3}^2 + \ParamRotMat_{2,3}^2 + \ParamRotMat_{3,3}^2 = 1$ 
    $\Leftrightarrow x_8^2 + x_{11}^2 + x_{14}^2 = x_5 x_{18}$\\
    $\Rightarrow 
    \QCQPStateVec^\top \QCQPConsMatrix_{10} \QCQPStateVec = \QCQPConsScalar_{10}$,
    \item $\ParamRotMat_{1,1} \ParamRotMat_{1,2} + \ParamRotMat_{2,1} \ParamRotMat_{2,2} + \ParamRotMat_{3,1} \ParamRotMat_{3,2} = 0$ \\
    $\Leftrightarrow x_6 x_7 + x_9 x_{10} + x_{12} x_{13} = 0$
    $\Rightarrow 
    \QCQPStateVec^\top \QCQPConsMatrix_{11} \QCQPStateVec = \QCQPConsScalar_{11}$,
    \item $\ParamRotMat_{1,1} \ParamRotMat_{1,3} + \ParamRotMat_{2,1} \ParamRotMat_{2,3} + \ParamRotMat_{3,1} \ParamRotMat_{3,3} = 0$ \\
    $\Leftrightarrow x_6 x_8 + x_9 x_{11} + x_{12} x_{14} = 0$
    $\Rightarrow 
    \QCQPStateVec^\top \QCQPConsMatrix_{12} \QCQPStateVec = \QCQPConsScalar_{12}$,
    \item $\ParamRotMat_{1,2} \ParamRotMat_{1,3} + \ParamRotMat_{2,2} \ParamRotMat_{2,3} + \ParamRotMat_{3,2} \ParamRotMat_{3,3} = 0$ \\
    $\Leftrightarrow x_7 x_8 + x_{10} x_{11} + x_{13} x_{14} = 0$
    $\Rightarrow 
    \QCQPStateVec^\top \QCQPConsMatrix_{13} \QCQPStateVec = \QCQPConsScalar_{13}$,
\end{itemize}
where
\begin{align*} 
    &\QCQPConsMatrix_8 = \sparse{[6,9,12,5],[6,9,12,18],[1,1,1,-1],18,18}, \\
    &\QCQPConsMatrix_9 = \sparse{[7,10,13,5],[7,10,13,18],[1,1,1,-1],18,18}, \\
    &\QCQPConsMatrix_{10} = \sparse{[8,11,14,5],[8,11,14,18],[1,1,1,-1],18,18}, \\
    &\QCQPConsMatrix_{11} = \sparse{[6,9,12],[7,10,13],[1,1,1],18,18}, \\
    &\QCQPConsMatrix_{12} = \sparse{[6,9,12],[8,11,14],[1,1,1],18,18}, \tag{\stepcounter{equation}\theequation}\\
    &\QCQPConsMatrix_{13} = \sparse{[7,10,13],[8,11,14],[1,1,1],18,18}, \\
    &\QCQPConsScalar_8 = 0, \;
    \QCQPConsScalar_9 = 0, \;
    \QCQPConsScalar_{10} = 0, \;
    \QCQPConsScalar_{11} = 0, \;
    \QCQPConsScalar_{12} = 0, \;
    \QCQPConsScalar_{13} = 0.
\end{align*}

Let $\est{\QCQPStateVec}$ be the solution obtained from solving the QCQP problem (\ref{eq:prob_QCQP}). 
Since $x_{18}$ should be positive as can be seen in Eq. (\ref{eq:x_state_vector}), we flip the sign of $\est{\QCQPStateVec}$ if ${\est{x}_{18}}$ is negative.
We can then recover the parameters of interest from $\est{\QCQPStateVec}$ as follows:
\begin{equation}
\begin{aligned} 
    &\est{\ParamTrans} = \est{\QCQPStateVec}_{1:3}, \quad
    \est{\ParamScale} = \sqrt{\est{\QCQPStateVec}_5}, \quad
    \est{\ParamRotMat} = 
    \frac{1}{\est{\ParamScale}}
    \begin{bmatrix}
        \est{\QCQPStateVec}_{6:8} &
        \est{\QCQPStateVec}_{9:11} &
        \est{\QCQPStateVec}_{12:14}
    \end{bmatrix}^\top.
\end{aligned}
\end{equation}

\subsubsection{Refinement with NLS} \label{subsec:UWBReloc}

Since running the QCQP estimation can be computationally costly, we only run it to obtain an initial guess $\{\est{\ParamTrans}_0,\est{\ParamRotMat}_0,\est{\ParamScale}_0\}$ and compute
\begin{equation}
    \est{\ParamRotAngle}_0
    = \arccos{\frac{\Tr{\est{\ParamRotMat}_0} - 1}{2}}, \;
    \est{\ParamRotVec}_0 = \frac{\est{\ParamRotAngle}_0}{2 \sin{\est{\ParamRotAngle}_0}}
    \begin{bmatrix}
        \est{\ParamRotMat}_{3,2} - \est{\ParamRotMat}_{2,3} \\
        \est{\ParamRotMat}_{1,3} - \est{\ParamRotMat}_{3,1} \\
        \est{\ParamRotMat}_{2,1} - \est{\ParamRotMat}_{1,2}
    \end{bmatrix}.
\end{equation}
In the identity rotation case, i.e. $\est{\ParamRotMat}_0 = \IdMat$, then $\est{\ParamRotAngle}_0 = 0$ and $\est{\ParamRotVec}_0$ can be chosen arbitrarily. In practice, if $\est{\ParamRotAngle}_0$ is smaller than a threshold, e.g. $1e^{-3}$, we do not update $\est{\ParamRotVec}_0$.
Then, with every new data point, we refine the estimates by solving the nonlinear least squares optimization problem:
\begin{equation} \label{eq:GAT_NLS}
\begin{aligned}
    \{\est{\ParamTrans},
    \est{\ParamRotVec},
    \est{\ParamScale} \}
    = &\argmin_{\ParamTrans,\ParamRotVec,\ParamScale}
    \sum\limits_{i=1}^{k}
    \left(
    \tilde{d}_i - \norm{\ParamTrans + \ParamScale \AlignedPos_i - \Wap}
    \right)^2. \\
\end{aligned}
\end{equation}

Let $\est{\textrm{CRLB}}$ be the CRLB computed with the estimated parameters $\est{\StateVector}$. For the $i$-th parameter ($i \in [1,\cdots,7]$), we measure $\est{\sigma}_{\StateVector_i} \coloneqq \sqrt{\est{\textrm{CRLB}}_{i,i}}$ as the  standard error of the estimate.
The overall uncertainty of the estimates is measured by $\max{\est{\sigma}_{\StateVector_i}}$ (i.e., the largest standard error of all estimates). We run the NLS estimation until $\max{\est{\sigma}_{\StateVector_i}}$ is sufficiently small then $\est{\StateVector}_i$ is fixed as the final value. The VO frame $\VOFrame$ and all internal data of visual SLAM are then transformed into the world frame $\WorldFrame$, and the next phase (VRO) begins. Conversely, if $\max{\est{\sigma}_{\StateVector_i}}$ is too large ($> 1000$ in our experiments), the trajectory configuration is considered singular with $\Theta_i$ as the unobservable parameter.

\subsection{Tightly-coupled Visual-Range Odometry (VRO)} \label{subsec:VRO}

\subsubsection{UWB-aided Bundle Adjustment (UBA)} \label{subsubsec:UBA}

The state vector and cost function used in the tracking, mapping and loop closing thread can be written in a general form as:
\begin{equation} \label{eq:cost_function_UBA}
\begin{aligned}
    &\mathbfcal{X}_1 
    = [
    \WCp_1^\top,\WCq_1^\top, \cdots
    \WCp_K^\top,\WCq_K^\top,
    \prescript{\mathcal{W}}{v_1}{\mathbf{p}}^\top,\cdots,
    \prescript{\mathcal{W}}{v_L}{\mathbf{p}}^\top
    ]^\top, \\
    &E_1 (\mathbfcal{X}_1) 
    = 
    \sum\limits_{i=1}^{K} \sum\limits_{l=1}^{L} {\mathbf{e}_{\mathbf{V}}^{i,l}}^\top 
    \mathbf{W}^v 
    {\mathbf{e}_{\mathbf{V}}^{i,l}} 
    + 
    \sum\limits_{i=1}^{K} \sum\limits_{n=1}^{N} 
    \VarR^{-1}\varrho(\abs{e_{\mathbf{R}}^{i,n}}^2),
\end{aligned}
\end{equation}
where $K$ is the number of KFs ($\mathcal{C}$) and $L$ is the number of visual landmarks ($v$) observed in the map. The main difference is that the tracking thread optimizes only the current camera pose ($K=1$), the local mapping thread optimizes a local window of KFs and map points, and the loop closure thread optimizes all KFs and mappoints (Full BA). The main novelty is the inclusion of the UWB ranging residual $e_{\mathbf{R}}^{i,n}$.
\begin{itemize}
    \item The reprojection residual $\mathbf{e}_{\mathbf{V}}^{i,l}$ is defined as:
    \begin{equation} \label{ch04:eqa:visual_residual}
        \mathbf{e}_{\mathbf{V}}^{i,l} = \mathbf{z}(i,l) - \hat{\mathbf{z}}(i,l),
    \end{equation}
    where $\mathbf{z}(i,l)$ is the measurement of the $l$-th visual landmark in the $i$-th camera frame and $\hat{\mathbf{z}}(i,l)$ is the predicted value of $\mathbf{z}(i,l)$. If the $l$-th landmark is not observed in the $i$-th frame then $\mathbf{e}_{\mathbf{V}}^{i,l} = 0$. The residuals are weighted by $\mathbf{W}^v$, which is the uniform information matrix for all visual measurements.
    \item The ranging measurement residual $e_{\mathbf{R}}^{i,n}$ is defined as:
    \begin{equation} \label{eq:uwb_res}
        e_{\mathbf{R}}^{i,n} = \tilde{d}_{i,n} - \norm{\WCp_i - \Wap},
    \end{equation}
    which includes the $i$-th KF and the $n$-th anchor. $e_{\mathbf{R}}^{i,n} = 0$ if no measurement to $\Wap$ is available. The Huber norm $\varrho(\cdot)$ is used to reduce the impact of any spurious UWB measurements.
\end{itemize}

\subsubsection{UWB-aided Pose Graph Optimization (UPGO)} \label{subsubsec:UPGO}

\begin{figure}[t]
\centering
\includegraphics[width=\linewidth]{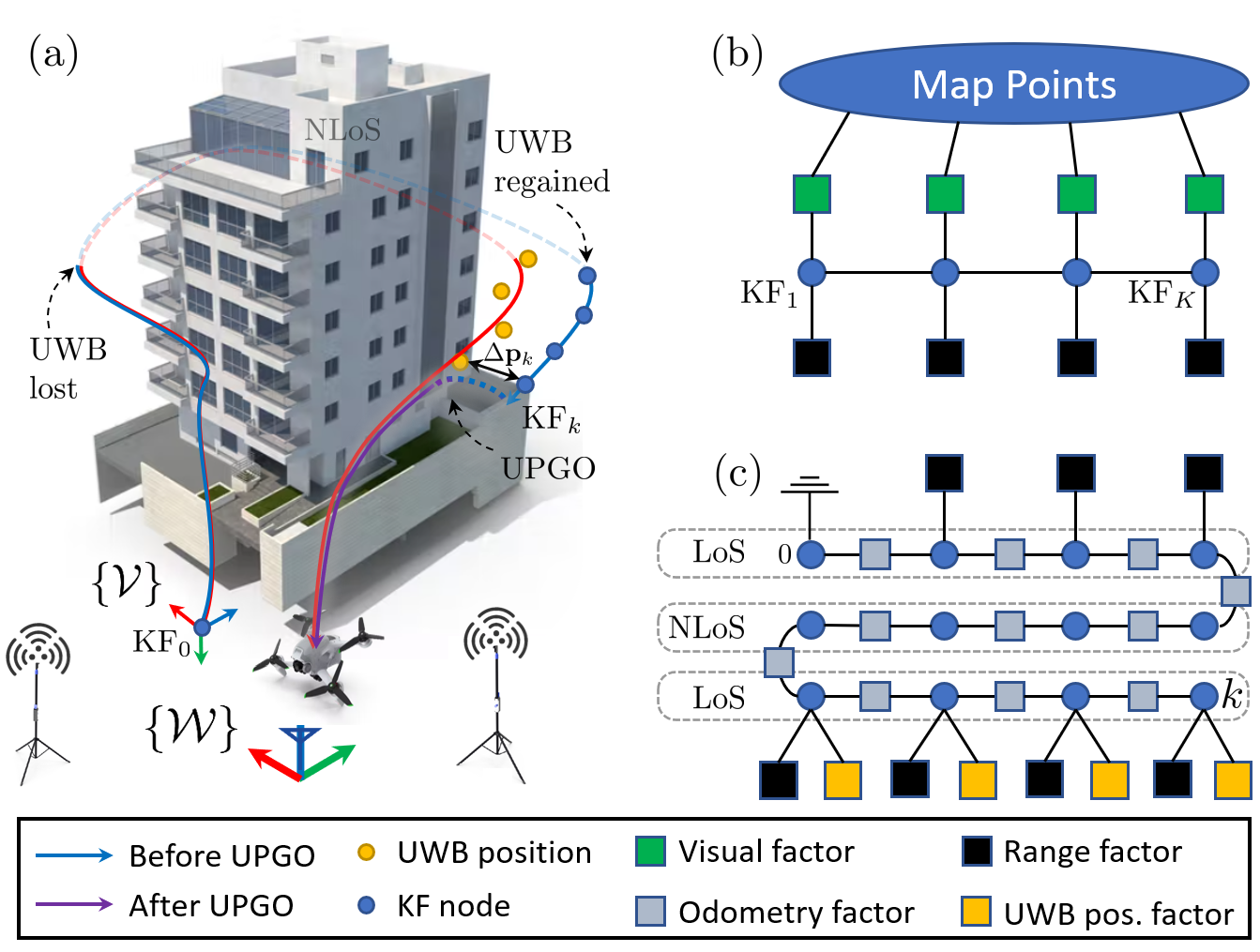}
    \caption{Illustration of a) the main idea for the UPGO module, b-c) factor graphs of the UBA and UPGO, respectively.}
    \label{fig:factorgraph}
\end{figure}

Fig. \ref{fig:factorgraph} visualizes the scenario and the factor graph of the UPGO.
While UBA can combine visual and ranging data to reduce the long-term drift, one issue that can occur in real operation is the intermittent loss of UWB signal. For example, the robot might move away from the area with LoS to UWB anchors to perform some tasks, then come back once finished. During the non-line-of-sight (NLoS) time, the camera will be the only sensor for localization and thus errors will accumulate over time. When the UWB sensor is back online, the UBA might not be able to correct the errors since the KFs are outside of the sliding window. As such, there would be a large and growing inconsistency between ranging and visual residuals, which can significantly affect the performance.
The UPGO module is designed to address such cases. In case a loop closure is detected during either of the following steps, we will abandon the task, wait until the full BA is finished and restart the whole process from step 1.

\paragraph*{Step 1: UWB-based drift detection}

As running UPGO can be computationally costly, we perform a check to see whether it is necessary to do so. At time instance $k$, if there are enough range measurements to find a unique position ($N \geq 3$ and assuming that the robot's height is positive), we estimate the UWB-only position $\prescript{\mathcal{W}}{\UWBAntenna}{\mathbf{p}_k}$ by solving the optimization:
\begin{equation} \label{eq:cost_uwb_only}
\begin{aligned}
    \prescript{\mathcal{W}}{\UWBAntenna}{\est{\mathbf{p}}_k}
    = &\argmin_{\prescript{\mathcal{W}}{\UWBAntenna}{\mathbf{p}}_k}
    \sum\limits_{n=1}^{N}
    \left(
    \tilde{d}_{k,n} - \norm{\prescript{\mathcal{W}}{\UWBAntenna}{\mathbf{p}}_k - \Wap}
    \right)^2,
\end{aligned}
\end{equation}
where $\UWBAntenna$ denotes the UWB antenna on the robot.
We then compute the translational offset between the positions estimated by only UWB and VR-SLAM as:
\begin{equation}
\begin{aligned}
    \Delta \est{\mathbf{p}}_k = \norm{\prescript{\mathcal{W}}{\mathcal{C}}{\est{\mathbf{p}}_k} - \prescript{\mathcal{W}}{\UWBAntenna}{\est{\mathbf{p}}_k}}.
\end{aligned}
\end{equation}

The idea is that while $\prescript{\mathcal{W}}{\UWBAntenna}{\est{\mathbf{p}}_k}$ might not be very accurate, it is independent of visual measurements and is not affected by drift. Hence, $\Delta \est{\mathbf{p}}_k$ can be used to measure the drift. We collect $P$ samples $\{\Delta \est{\mathbf{p}}_i\}_{i=k-P+1,\cdots,k}$ in a sliding window and compute the mean value. If the mean value is higher than a threshold (e.g., $1\si{m}$), the UPGO will begin. 

\paragraph*{Step 2: UPGO Formulation}

The variables to be optimized are the KF's poses, denoted as $\mathcal{T}_i \coloneqq \prescript{\mathcal{W}}{\mathcal{C}}{\mathbf{T}}_i$ ($i\in[0,\cdots,k]$). The cost function for UPGO is defined as:
\begin{equation}
\begin{aligned}
    \min_{\mathcal{T}_1,\cdots,\mathcal{T}_k}
    \sum\limits_{i=0}^{k} \sum\limits_{n=1}^{N} 
    \abs{e_{\mathbf{R}}^{i,n}}^2
    +
    \sum\limits_{i=0}^{k}
    \norm{e_{\mathbf{O}}^i}^2
    +
    \sum\limits_{i=k-P+1}^{k}
    \varrho(\norm{e_{\mathbf{P}}^i}^2),
\end{aligned}
\end{equation}
 where we employ three types of factors:
\begin{itemize}
    \item Range factor $e_{\mathbf{R}}^{i,n}$, which provides the distance constraint between the KF' positions and the anchors' positions;
    \item Odometry factor $e_{\mathbf{O}}^{i}$, which gives the relative transformation constraint between the $i-1$-th and $i$-th KFs;
    \item Position factors $e_{\mathbf{P}}^{i}$, which directly link the KFs' positions with the UWB positions $\prescript{\mathcal{W}}{\UWBAntenna}{\est{\mathbf{p}}_i}$, and has already been computed in the drift detection step so can be used without any extra computation cost.
\end{itemize}

$e_{\mathbf{R}}^{i,n}$ is defined in Eq. (\ref{eq:uwb_res}) while $e_{\mathbf{O}}^{i}$ and $e_{\mathbf{P}}^{i}$ are:
\begin{equation}
    e_{\mathbf{O}}^{i} = (\mathcal{T}_i^{-1} \mathcal{T}_{i-1}) \ominus (\est{\mathcal{T}}_i^{-1} \est{\mathcal{T}}_{i-1}), \quad
    e_{\mathbf{P}}^{i} = \prescript{\mathcal{W}}{\mathcal{C}}{\mathbf{p}}_i - \prescript{\mathcal{W}}{\UWBAntenna}{\est{\mathbf{p}}_i},
\end{equation}
where 
$\prescript{\mathcal{A}}{}{\mathbf{T}} \ominus \prescript{\mathcal{B}}{}{\mathbf{T}} =
\begin{bmatrix}
    \prescript{\mathcal{A}}{}{\mathbf{p}} - \prescript{\mathcal{B}}{}{\mathbf{p}} \\
    \prescript{\mathcal{A}}{}{\mathbf{e}} - \prescript{\mathcal{B}}{}{\mathbf{e}}
\end{bmatrix} \in \R^{6\times1}$,
$\prescript{\mathcal{A}}{}{\mathbf{e}}$ and $\prescript{\mathcal{B}}{}{\mathbf{e}}$ are the vectors of the Euler angles (e.g., roll-pitch-yaw) representing the rotation matrices $\prescript{\mathcal{A}}{}{\mathbf{R}}$ and $\prescript{\mathcal{B}}{}{\mathbf{R}}$.
Note that if there are not enough range factors that can provide a position for the first KF (i.e., $N < 3$ for $i=0$), then the first vertex is fixed since it serves as the origin point. After UPGO is done, we run a full BA but fix all KF vertexes and only optimize the mappoints, as a way to update the map based on the result of UPGO.

\subsubsection{UWB-aided Visual Relocalization (UVR)} 

\textit{Visual loop closure verification:} if a new loop closure candidate is found using camera images, the new camera pose $\prescript{\mathcal{W}}{\mathcal{C}}{\check{\mathbf{p}}_k}$ is accepted if the predicted ranges are comparable to the latest real measurements, i.e. for a pre-defined limit $\epsilon_r$:
\begin{equation}
    \abs{\norm{\prescript{\mathcal{W}}{\mathcal{C}}{\check{\mathbf{p}}_k} - \Wap} - \RangeKNnoisy} \leq \epsilon_r \; \forall n.
\end{equation}
Otherwise, we discard this loop closure candidate.

\textit{Re-initializing in the world frame:} 
In the event that tracking got lost at time $k' < k$, we save the last camera pose $\prescript{\mathcal{W}}{\mathcal{C}}{\est{\mathbf{p}}_{k'}}$, restart the VO frontend, then attempt to relocalize using both visual and UWB data. If the system still cannot relocalize with images after we have collected enough odometry and range data, we re-solve the NLS problem (\ref{eq:GAT_NLS}) using $\prescript{\mathcal{W}}{\mathcal{C}}{\est{\mathbf{p}}_{k'}}$ as the initial guess for the translation and rotation parameters, while the initial guess for the scale factor is the last result $\est{\ParamScale}$. However, if $\max{\est{\sigma}_{\StateVector_k}}$ cannot get smaller than the required threshold after a number of runs, we switch back to solving the QCQP problem (\ref{eq:prob_QCQP}) to obtain a new initial guess.

\section{Experimental Results} \label{sec:ExpResults}

In this section, we evaluate the performance of the VR-SLAM system as well as its components in different settings, starting from simulation to public dataset and finally challenging real-life experiments.
A video of the simulations and experiments can be found at \url{https://youtu.be/rd4kf25gxPc}. 

\subsection{Simulation}

In this section, we aim to evaluate the GAT estimation module in the most general way. We simulate a micro aerial vehicle (MAV) flying in a $5\si{m} \times 5\si{m}$ area, with four UWB anchors located at $[0,0,0]$, $[5,0,1]$, $[0,5,2]$ and $[5,5,3]$. Let $R_m$ be the maximum motion radius of the MAV, i.e. the MAV does not move further than $R_m$ meter away from its starting position at any point during the simulation. For one value of $R_m$, we perform $100$ Monte-Carlo simulations with the robot starting position (within the $5\si{m} \times 5\si{m}$ area), the MAV's trajectory and the true affine transformation $\ParamGAT$ all randomized. The UWB and odometry data are corrupted by noises with standard deviations of $\VarR = 0.1\si{m}$ and $\sigma_o = 0.001\si{m}$, respectively. This process allows our results to be more statistically significant.
 
We measure the error of the estimated translation and scale factor as $e_t = \norm{\est{\ParamTrans} - \ParamTrans}$ and $e_s = \abs{\est{\ParamScale} - \ParamScale}$, respectively. For the rotation, we first convert the estimated parameters from each algorithm to their rotation matrix form $\est{\ParamRotMat}$, then compute the angle of the difference rotation as the error:
\begin{equation}
    e_R = \arccos{\frac{\Tr{\ParamRotMat^\top \est{\ParamRotMat}} - 1}{2}}.
\end{equation}

\begin{figure}[t]
\centering
\includegraphics[width=\linewidth]{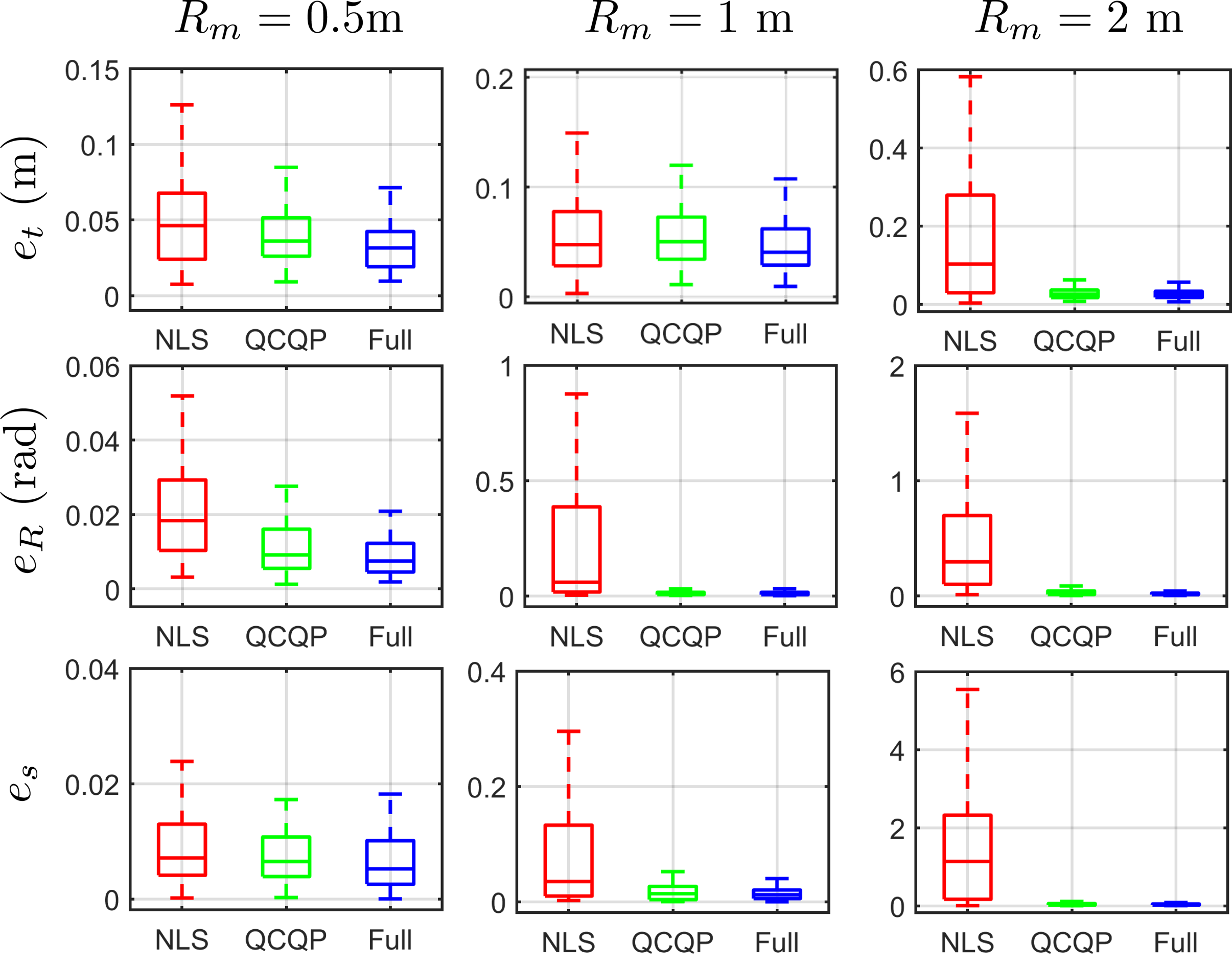}
    \caption{Results in simulation for the GAT module.}
    \label{fig:sim_results}
\end{figure}

Fig. \ref{fig:sim_results} shows the results of the NLS solver (\ref{eq:GAT_NLS}) without any initial guess, the QCQP solver (\ref{eq:prob_QCQP}) and the full system (first solve the problem with QCQP, then refine the solution with NLS).
Firstly, it can be seen that overall, the QCQP method is indeed more accurate than NLS and refining the QCQP's solution with NLS can marginally improve the result. Secondly, the larger the $R_m$ is, the smaller the errors become. This observation agrees with our theoretical analysis: in general, if the robot's motion is more spatially spread out, the relative angles between the sensing position would increase, which in turn increases the volume of $\PartialFs_j$, $\DetSubRot_i$, and $\DetSubTrans_i$ (Eq. \ref{eq:detFPhys}) or equivalently the value of $\detF$ and conversely lower the error covariance. Overall, our full solution (initialize with QCQP, then refine with NLS) provides the most accurate results, even in challenging cases where the robot's motion is confined to a small area.

\subsection{Public Dataset}

We evaluate the VR-SLAM system with the NTU-VIRAL dataset \cite{nguyen2021viraldataset}, which includes a MAV flying in various outdoor and indoor environments and $N=3$ UWB anchors. We use the left camera, which provides image at $10\si{Hz}$, and the UWB ranging data which comes at $68\si{Hz}$ in our algorithm. The 3-DoF position ground truth is obtained by a Leica MS60 MultiStation system at $20\si{Hz}$. For all sequences, before the GAT estimation begins we collect the VO's odometry and UWB ranging data until the MAV's motion has covered all three axes to avoid most singular cases. 
In Sect. \ref{subsec:VIRAL_GAT_result}, we compare the GAT estimation results of all methods after a fixed amount of time (e.g., $20\si{s}$ after the first estimation). On the other hand, in Sect. \ref{subsec:VIRAL_Odom_result} we conclude the GAT phase when $\max{\est{\sigma}_{\StateVector_k}}$ is considered small enough ($< 0.1$ in the experiments), then let the VRO phase begins. 

\subsubsection{GAT estimation accuracy} \label{subsec:VIRAL_GAT_result}

\begin{table}[t]
\centering
\begin{adjustbox}{width=\columnwidth}
    \begin{tabular}[t]{c||ccc|ccc}
    \toprule
    \multirow{2}{*}{\tb{Sequence}}
    & \multicolumn{3}{c|}{NLS only}
    & \multicolumn{3}{c}{QCQP + NLS} \\ 
    \cline{2-4} \cline{5-7}
    & $e_t$ & $e_R$ & $e_s$ & $e_t$ & $e_R$ & $e_s$\\
    \hline
    {eee\_01}
            & 0.418 & 0.148 & 0.115 
            & 0.122 & 0.022 & 0.035 \\
    {eee\_02}
            & 1.175 & 0.837 & 0.196 
            & 0.676 & 0.033 & 0.079 \\
    {eee\_03}
            & 0.553 & 0.033 & 0.123 
            & 0.312 & 0.008 & 0.113 \\
    {nya\_01}
            & 0.857 & 0.087 & 0.448 
            & 0.214 & 0.025 & 0.365 \\
    {nya\_02}
            & 2.332 & 0.148 & 0.770 
            & 0.205 & 0.038 & 0.336 \\
    {nya\_03}
            & 1.365 & 2.466 & 0.553 
            & 0.442 & 0.038 & 0.452 \\
    \bottomrule
    \end{tabular}
\end{adjustbox}
\caption{GAT estimation errors in the NTU VIRAL dataset. $e_t$ and $e_R$ are measured in $\si{m}$ and $rad$, respectively.} \label{table:GAT_NTU_VIRAL}
\end{table}

\begin{figure}[t]
\centering
\includegraphics[width=\linewidth]{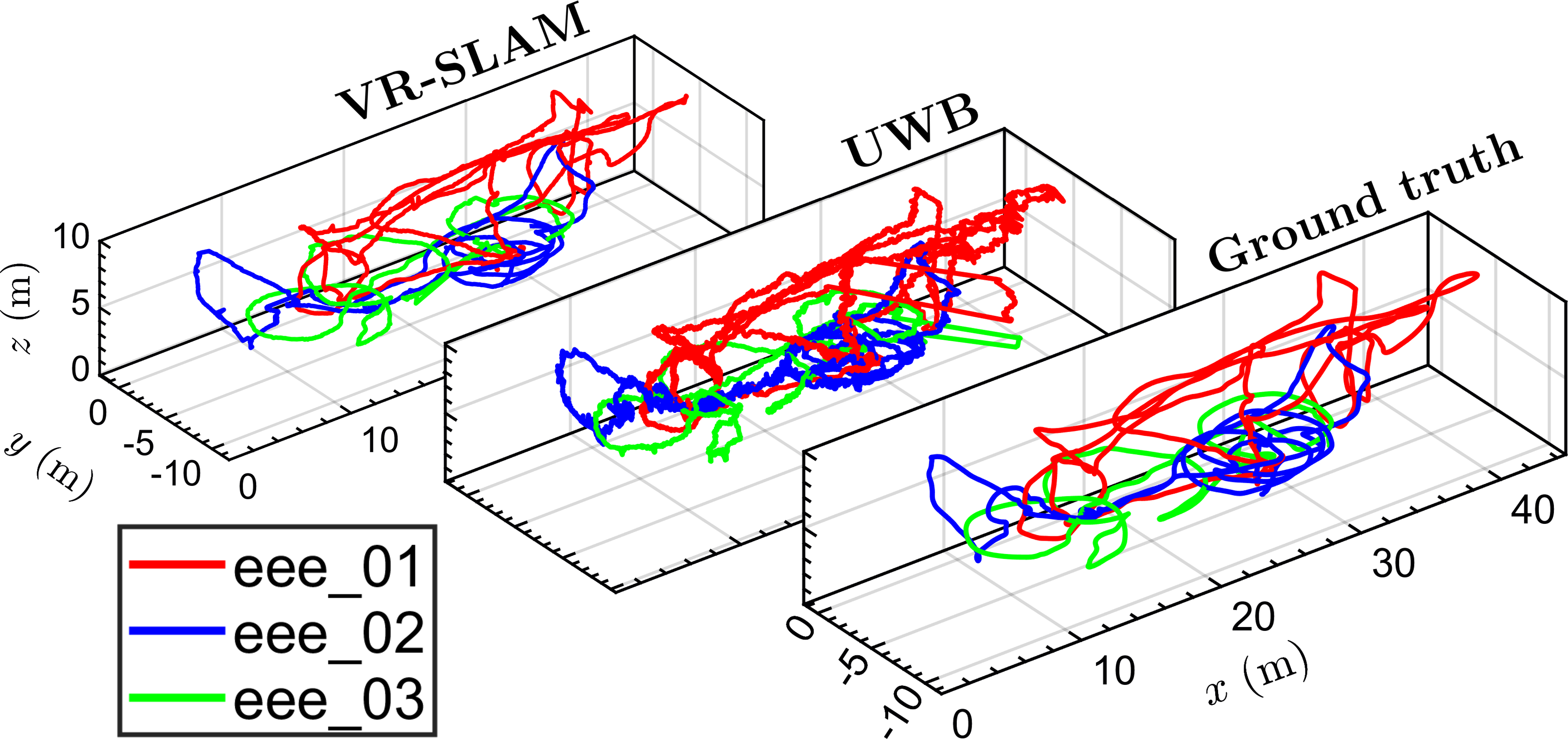}
    \caption{The estimated and ground truth trajectories in all \textbf{eee\_01-03} sequences.}
    \label{fig:viral_eee_aligned}
\end{figure}

Table \ref{table:GAT_NTU_VIRAL} shows the estimation errors for the GAT problem. No initial guess was provided for either the NLS method or our full solution. As expected, the NLS method's results are much less accurate than the combined approach, mainly thanks to the QCQP module's ability to obtain a good first estimate.
Fig. \ref{fig:viral_eee_aligned} illustrates all trajectories estimated by our solution, UWB-only, and the ground truth system. Note that our results are obtained without any post-processing steps, i.e. these trajectories are automatically scaled and aligned in the world frame while running each sequence. Monocular VSLAM was not able to combine the trajectories in the same manner since there are no loop closures across all sequences. It can be seen that while UWB-only's result also closely resembles the ground truth, the trajectories are much less smooth and more importantly, there is no map of the environment to be obtained. On the other hand, our system can produce not only much smoother trajectories that also closely follow the ground truth, but also a map of visual landmarks of the environments.

\begin{figure}[t]
\centering
\includegraphics[width=\linewidth]{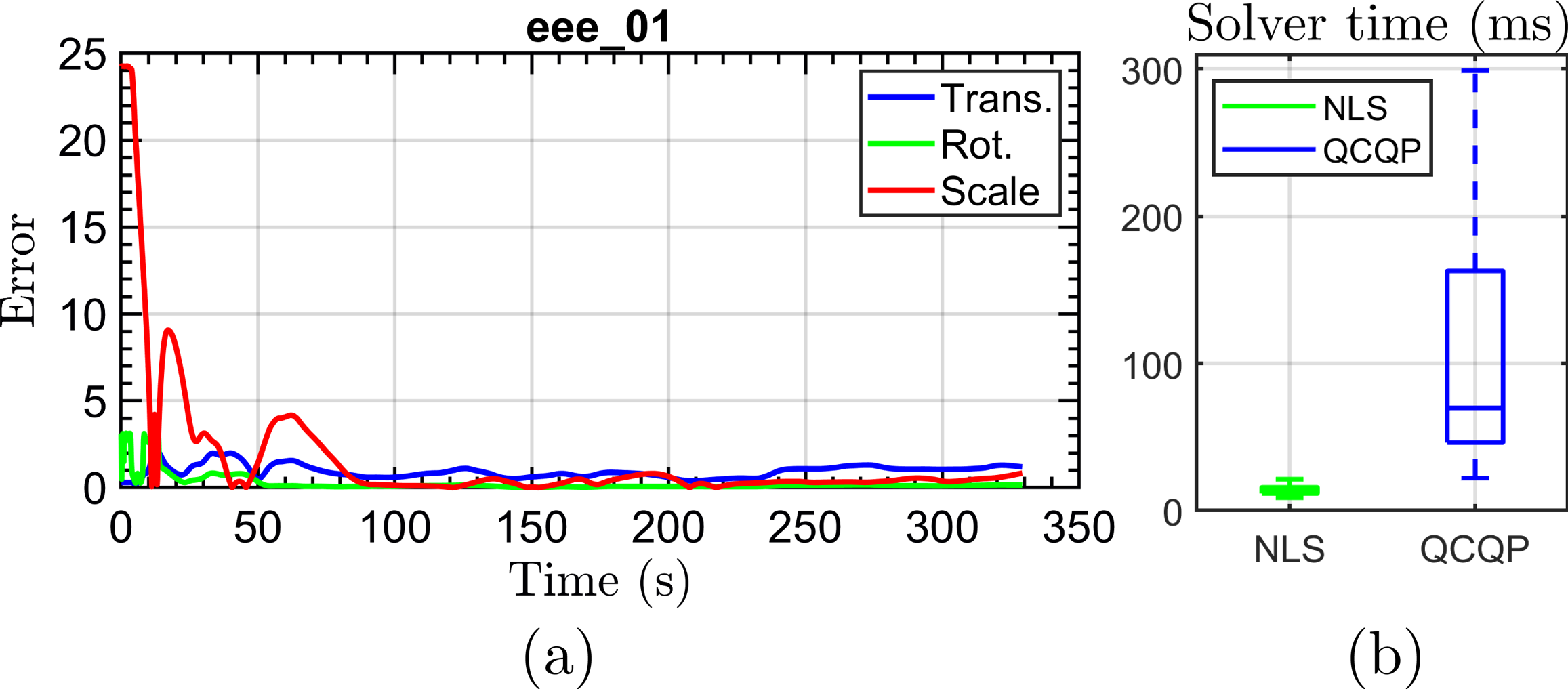}
    \caption{Left: The GAT estimation error in the eee\_01 sequence. Right: Comparison of the solver time between the NLS and QCQP methods.}
    \label{fig:compare_NLS_vs_QCQP}
\end{figure}

Fig. \ref{fig:compare_NLS_vs_QCQP}a illustrates the errors for the estimates of the translation, rotation and scale if we run the GAT module continuously until the end of the \textbf{eee\_01} sequence. It can be seen that the QCQP estimator was able to find a good initial result, as evident by the sharp drop in the errors at the beginning of the sequence. The estimates are then further refined as more data are obtained.
Although the QCQP method is generally more accurate than the NLS method, it is much slower in terms of computing time. Fig. \ref{fig:compare_NLS_vs_QCQP}b shows the average solver time in all of our experiments, which clearly shows the superiority of the NLS solution in this regard.

\subsubsection{Odometry accuracy} \label{subsec:VIRAL_Odom_result}

\begin{table}[t]
\centering
\begin{adjustbox}{width=\columnwidth}
    \begin{tabular}[t]{c||cccc|cc}
    \toprule
    \multirow{2}{*}{\tb{Sequence}}
    & UWB    
    & VSLAM   
    & VINS-
    & VIRO-
    &\multicolumn{2}{c}{VR-SLAM}\\ 
    \cline{6-7}
    & (RT) & (OT) & Fusion & FEJ & RT & OT\\
    \hline
    {eee\_01}
            & 0.507 & \ul{0.338} 
            & 0.608 & 0.479 
            & 0.357 & \tb{0.157} \\
    {eee\_02}
            & 0.445 & 0.875 
            & 0.506 & \ul{0.340} 
            & 0.353 & \tb{0.144} \\
    {eee\_03}
            & 0.557 & 0.401 
            & 0.494 & \ul{0.348}
            & 0.435 & \tb{0.127} \\
    {nya\_01}
            & 0.425 & \ul{0.333} 
            & 0.397 & 0.505 
            & 0.374 & \tb{0.151} \\
    {nya\_02}
            & 0.504 & 0.550 
            & 0.424 & \tb{0.202} 
            & 0.390 & \ul{0.238} \\
    {nya\_03}
            & 0.723 & 0.488 
            & 0.787 & \ul{0.290} 
            & 0.302 & \tb{0.166} \\
    \bottomrule
    \end{tabular}
\end{adjustbox}
\caption{ATE (m) of different methods in NTU VIRAL dataset. RT and OT stand for real-time and optimized trajectory, respectively. The best and second best results are in \tb{bold} and \ul{underlined}, respectively.} \label{table:ATE_NTU_VIRAL}
\end{table}

Table \ref{table:ATE_NTU_VIRAL} shows the absolute trajectory error (ATE, in $\si{m}$) of UWB-only \cite{wang2017ultra}, monocular VSLAM \cite{mur2017orb} and our VR-SLAM methods. We report the results for the real-time (RT) position estimates and the final optimized trajectory (OT), i.e., all KF poses after they are corrected by visual loop closure and UPGO modules. We also include the RT results of VINS-Fusion \cite{qin2018vins}, a visual-inertial odometry method, and VIRO-FEJ \cite{shenhan2022viro}, a visual-inertial-range odometry solution (the OT results are not available).
From Table \ref{table:ATE_NTU_VIRAL}, it can be seen that our OT results are the most accurate solution, while UWB-only localization is the least. The UWB-only localization results are stable across the dataset, which can be an advantage in terms of reliability. The RT results of VR-SLAM and VIRO-FEJ are comparable, with VIRO-FEJ being marginally better.

\begin{figure}[t]
\centering
\includegraphics[width=\linewidth]{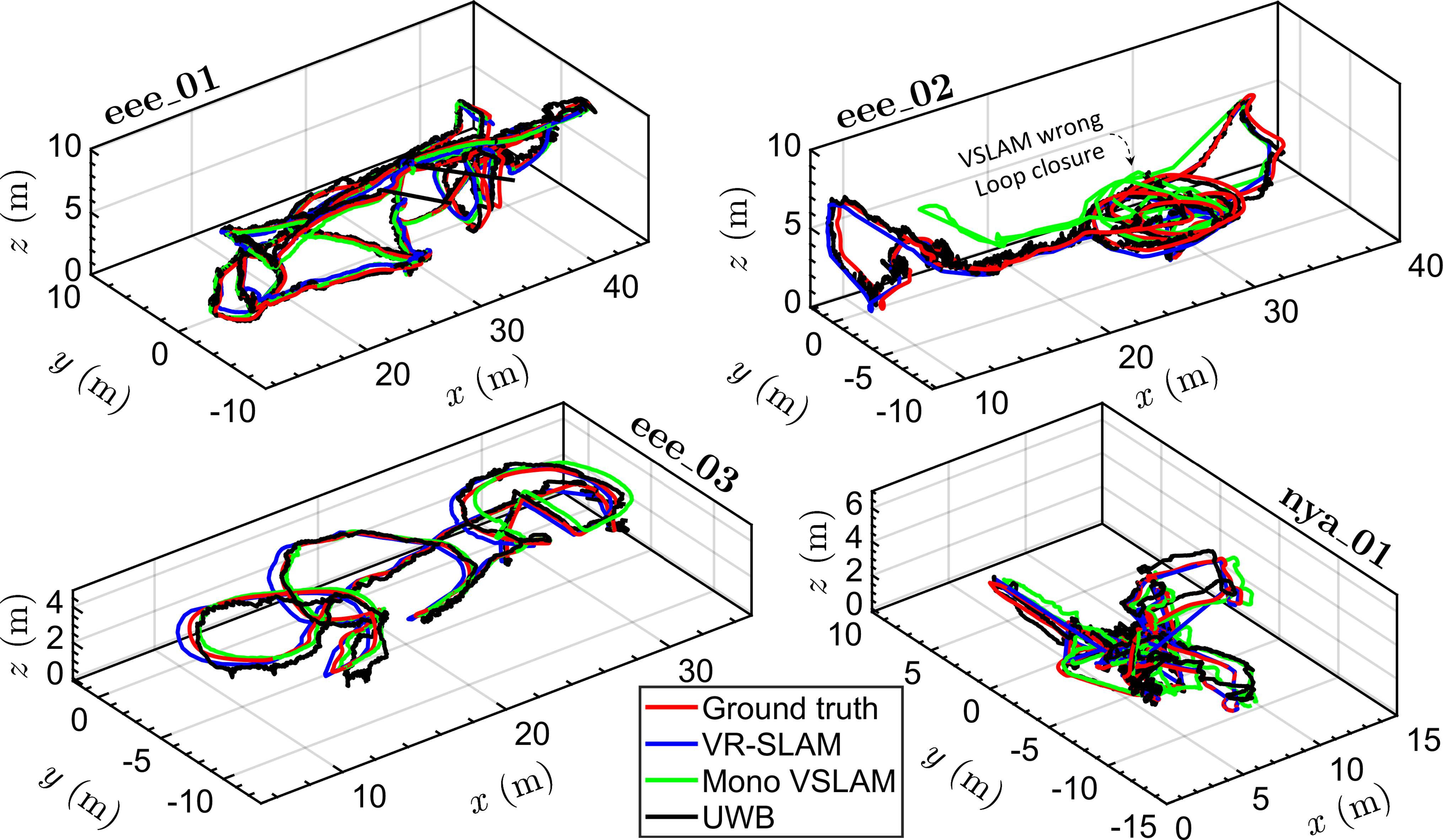}
    \caption{Overview of the estimated trajectories with NTU VIRAL dataset.}
    \label{fig:viral_traj_overview}
\end{figure}

\begin{figure}[t]
\centering
\includegraphics[width=\linewidth]{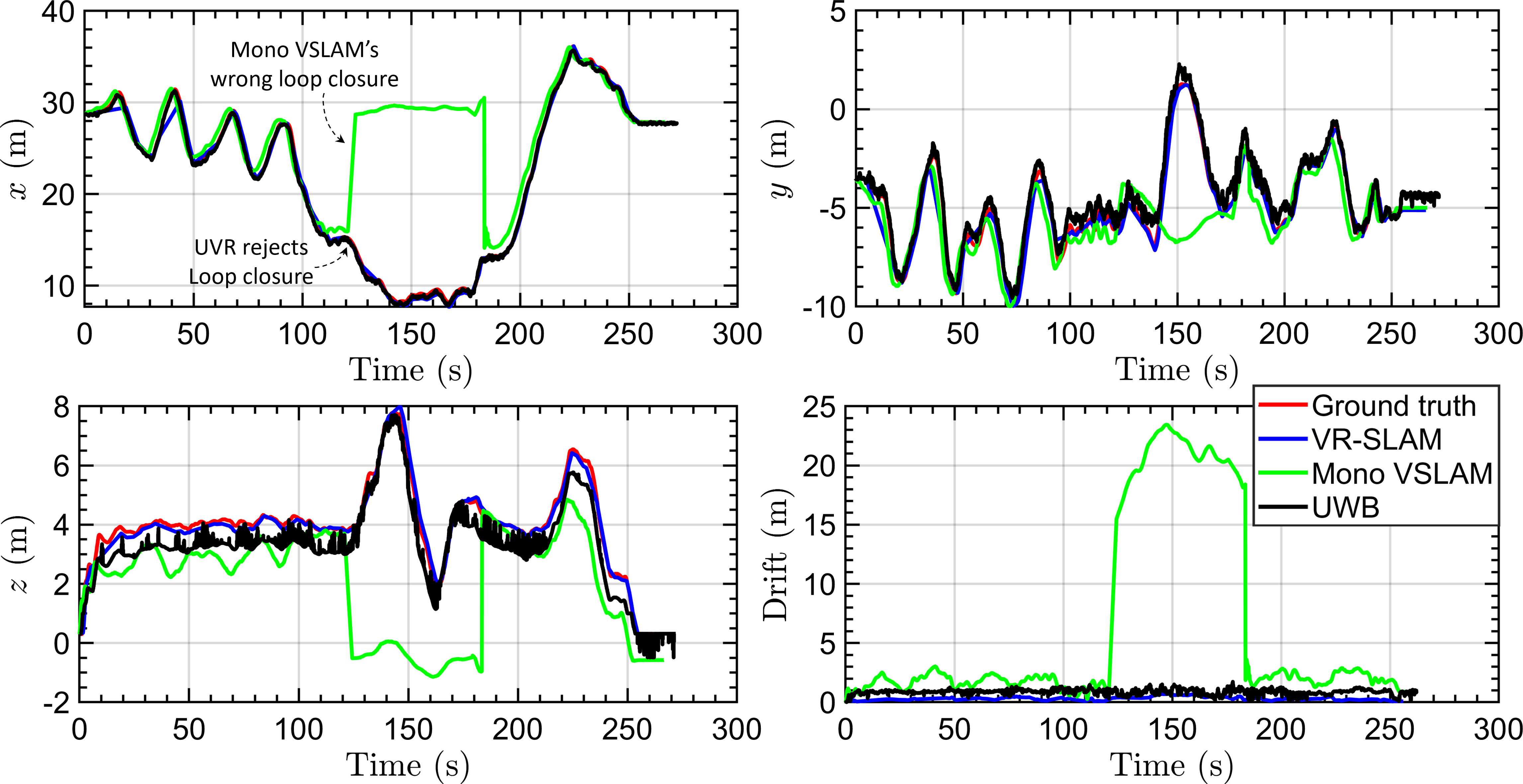}
    \caption{Estimated position and drift in \textbf{eee\_02} sequence.}
    \label{fig:eee02_xyzdrift}
\end{figure}

However, the ATE metric is computed over estimated poses. It does not give insights on the advantages of our methods and the shortcomings of the camera/UWB-only methods. Hence, we look further into the estimated trajectories for more analysis.
Fig. \ref{fig:viral_traj_overview} shows the 3D-aligned trajectories in some of the sequences with varying levels of difficulty. 
In the sequences with smooth, low-speed motions and no large rotations, all methods were able to successfully complete these sequences. 
In \textbf{eee\_02} sequence, the MAV faced a building with repeated structure, both horizontally and vertically. During the operation, Mono VSLAM mistakenly performed loop closures on two unrelated but visually similar locations, which consequently lead to significant errors and drift as shown in Fig. \ref{fig:eee02_xyzdrift}. Only at a much later point during the mission, Mono VSLAM was able to rectify this error since it detected a true loop closure. As a result, the ATE of Mono VSLAM increased significantly in this sequence. In contrast, VR-SLAM correctly rejected these loop closure candidates and did not suffer from these setbacks.

\subsection{Real-life Experiments}

\begin{figure}[t]
\centering
\includegraphics[width=\linewidth]{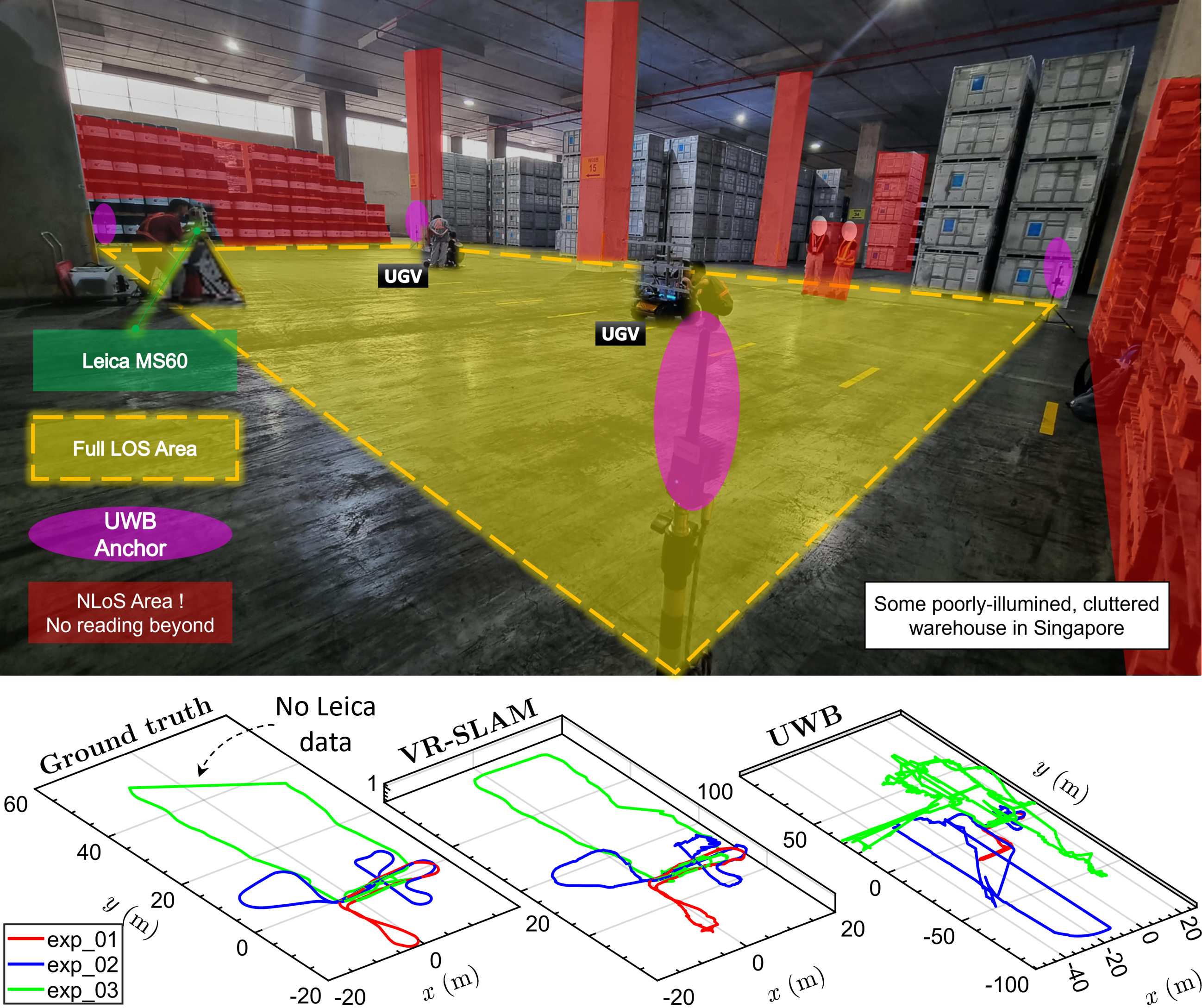}
    \caption{Top: The UGV and the setup in our real-life experiments. Bottom: The true and estimated trajectories.}
    \label{fig:real_life_setup}
\end{figure}

\begin{figure}[t]
\centering
\includegraphics[width=\linewidth]{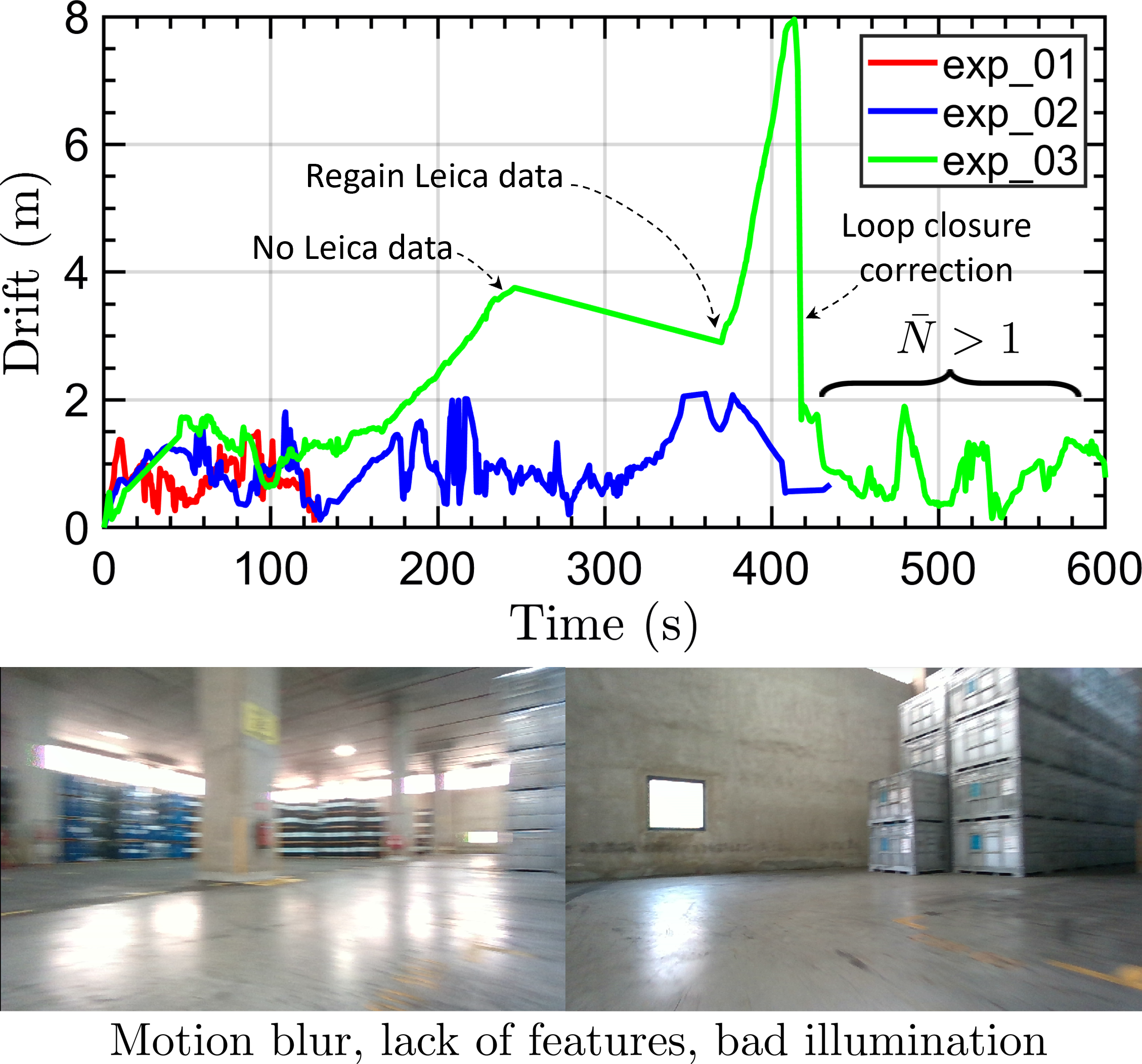}
    \caption{Top: Translation drifts of VR-SLAM in the experiments. Bottom: Examples of the failure cases of monocular VSLAM in our real-life experiments.}
    \label{fig:real_life_drifts}
\end{figure}

We conduct real-life experiments with an unmanned ground vehicle (UGV) to further evaluate the performance of VR-SLAM under NLoS conditions. The UGV is equipped with a Realsense D455 camera\footnote{\url{https://www.intelrealsense.com/depth-camera-d455/}} and a Nooploop UWB sensor \footnote{\url{https://www.nooploop.com/en/}}. We use a Leica MS60 MultiStation to collect the ground truth positions. The experiments are conducted in a large warehouse (Fig. \ref{fig:real_life_setup}). The UGV will start and end the mission in an $8\si{m} \times 18\si{m}$ area covered by $N=4$ UWB anchors placed at each corner. During the mission, it will then move outside to other areas that might not have clear LoS to all of the anchors (limited LoS area), or none at all (NLoS area). We design the trajectory such that the typical number of UWB anchors with LoS ($\bar{N}$) is the same throughout most of the sequence. The monocular VSLAM method was unable to complete any of the sequences and hence is not included in the comparison. Fig. \ref{fig:real_life_drifts} illustrates some challenging cases where VSLAM usually fails since it cannot consistently track the visual features.

\begin{table}[t]
\centering
\begin{adjustbox}{width=\columnwidth}
    \begin{tabular}[t]{cc||ccc}
    \toprule
    \multicolumn{2}{c||}{\tb{Sequence}}
    & UWB
    &\multicolumn{2}{c}{VR-SLAM} \\
    \cline{1-2}\cline{4-5}
    Name & Details & RT & RT & OT \\
    \hline
    {exp\_01} & 82.4$\si{m}$, 128$\si{s}$, $\bar{N}=3$ 
            & 1.72
            & \ul{1.12} & \tb{0.45} \\
    {exp\_02} & 129.2$\si{m}$, 421$\si{s}$, $\bar{N}=2$ 
            & 4.15 
            & \ul{1.26} & \tb{0.70} \\
    {exp\_03} & 207.8$\si{m}$, 600$\si{s}$, $\bar{N}=1$ 
            & 7.48 
            & \ul{3.51} & \tb{1.98} \\
    \bottomrule
    \end{tabular}
\end{adjustbox}
\caption{ATE (m) results in real-life experiments. Besides $\bar{N}$, the details for each sequence include the mission's total length ($\si{m}$) and time ($\si{s}$). The best and second best results are in \tb{bold} and \ul{underlined}, respectively. Monocular VSLAM method often fails in challenging cases as shown in Fig. \ref{fig:real_life_drifts}.} \label{table:ATE_real_exps}
\end{table}

Table \ref{table:ATE_real_exps} shows the ATE results of the UWB-only and VR-SLAM methods in three experiments. It is clear that overall, our VR-SLAM system outperforms both camera-only (could not finish the experiment) and UWB-only (finish but with large errors) solutions. As expected, with fewer anchors in LoS (smaller $\bar{N}$), the errors will increase for all methods, especially for UWB-only localization. 
Fig. \ref{fig:real_life_drifts} shows the translational drifts of VR-SLAM's estimates in the experiments. As can be seen, when $\bar{N} = 2$ or $3$ (exp\_01 and 02), the drifts can still be corrected and kept low throughout the mission. In exp\_03 ($\bar{N} = 1$), the drift still keeps growing over time and only a visual loop closure was able to reduce the error substantially. As such, although VR-SLAM is able to diminish the drift, the number of anchors in LoS is a major factor in the performance.

\section{Conclusion} \label{sec:Conclusions}

In this article, we present a SLAM framework, so-called VR-SLAM, which combines measurements from a monocular camera and UWB anchors to obtain accurate, drift-free and global results. We divide this sensor fusion problem into different sub-problems, specifically GAT and VRO, then loosely or tightly fuse visual and ranging data. We provide a detailed theoretical analysis for the GAT problem, including the derivation of the FIM, CRLB, and description of the singular configurations for each parameter. The accuracy and effectiveness of the proposed components are verified in simulations and experiments with a public dataset and a self-collected dataset. The results show that VR-SLAM can not only outperform both camera and UWB in all cases, but also combine their advantages and compensate for their weaknesses in a single package. As such, VR-SLAM holds great potential for many real-world applications that require a flexible, consistent, and reliable solution.
In the future, we aim to focus on improving the system's performance in limited LoS or NLoS conditions, as well as exploiting the semantic information from camera images, such as detecting static or moving objects, to enhance the robustness in dynamic environments.

\section*{Acknowledgements}
We would like to thank Mr. Xu Kuan and Mr. Yin Pengyu for their helps to conduct the real-life experiments.

\begin{appendices}

\section{}\label{appendix:derivatives}

In this section, let $\VCp_k = [\VCxyz_x, \VCxyz_y, \VCxyz_z]^\top$, $\Wap = [a_x, a_y, a_z]^\top$, $\ParamRotAxis = [\ParamRotVecxyz_x, \ParamRotVecxyz_y, \ParamRotVecxyz_z]^\top$. Denote $\UnitX$,$\UnitY$,$\UnitZ$ as the unit vectors along the 3 axes in $\R^3$.
We drop the $k$ and $n$ subscripts to improve the clarity of the equations where they are not important for the analysis.
Given that $\ParamRotMat = \cos{\ParamRotAngle} \; \IdMat_{3\times3} + \sin{\ParamRotAngle} \sksym{\ParamRotAxis} + (1 - \cos{\ParamRotAngle}) \ParamRotAxis \ParamRotAxis^\top$, 
we can expand $\Cap_\NumData$ as follows:
\begin{align*}
    &\Cap_\NumData = \ParamTrans + \ParamScale \AlignedPos_\NumData - \Wap \\
    &\mbox{\footnotesize $\displaystyle
    = \ParamTrans + (\ParamScale \cos{\ParamRotAngle}) \VCp_k + (\ParamScale \sin{\ParamRotAngle}) (\ParamRotAxis \cross \VCp_k) + \ParamScale (1 - \cos{\ParamRotAngle}) (\ParamRotAxis \cdot \VCp_k) \ParamRotAxis - \Wap
    $} \\
    &\mbox{\scriptsize $\displaystyle
    = 
    \begin{bmatrix}
        (\ParamScale \cos{\ParamRotAngle}) \VCxyz_x 
        + (\ParamScale \sin{\ParamRotAngle}) (\ParamRotVecxyz_y \VCxyz_z - \ParamRotVecxyz_z \VCxyz_y) 
        + \ParamScale (1 - \cos{\ParamRotAngle}) (\ParamRotAxis \cdot \VCp_k) \ParamRotVecxyz_x 
        + t_x - a_x \\
        (\ParamScale \cos{\ParamRotAngle}) \VCxyz_y 
        + (\ParamScale \sin{\ParamRotAngle}) (\ParamRotVecxyz_z \VCxyz_x - \ParamRotVecxyz_x \VCxyz_z) 
        + \ParamScale (1 - \cos{\ParamRotAngle}) (\ParamRotAxis \cdot \VCp_k) \ParamRotVecxyz_y 
        + t_y - a_y \\
        (\ParamScale \cos{\ParamRotAngle}) \VCxyz_z 
        + (\ParamScale \sin{\ParamRotAngle}) (\ParamRotVecxyz_x \VCxyz_y - \ParamRotVecxyz_y \VCxyz_x) 
        + \ParamScale (1 - \cos{\ParamRotAngle}) (\ParamRotAxis \cdot \VCp_k) \ParamRotVecxyz_z 
        + t_z - a_z 
    \end{bmatrix}
    $} \\
    &= 
    \begin{bmatrix}
        \rho_x & \rho_y & \rho_z
    \end{bmatrix}^\top.
\end{align*}

Next, we compute the derivatives of $f_k = \RangeKNtrue = \sqrt{\rho_x^2 + \rho_y^2 + \rho_z^2}$ with respect to the elements of $\StateVector$.

\subsection*{A.1) Derivative with respect to \texorpdfstring{$\ParamTrans$}{TEXT}}

The derivative of $f_k$ with respect to $t_x$ is:
\begin{equation}
\begin{aligned}
    \frac{\partial f_k}{\partial t_x} 
    = \frac{1}{2 \RangeKNtrue} 
    \Big(
    2 \rho_x \overbrace{\frac{\partial \rho_x}{\partial t_x}}^{=1} +
    2 \rho_y \overbrace{\frac{\partial \rho_y}{\partial t_x}}^{=0} +
    2 \rho_z \overbrace{\frac{\partial \rho_z}{\partial t_x}}^{=0}
    \Big)
    = \frac{\rho_x}{\RangeKNtrue}.
\end{aligned}
\end{equation}
Similarly:
\begin{equation}
\begin{aligned}
    \frac{\partial f_k}{\partial t_y} = \frac{\rho_y}{\RangeKNtrue}, \quad
    \frac{\partial f_k}{\partial t_z} = \frac{\rho_z}{\RangeKNtrue}.
\end{aligned}
\end{equation}
Note that $\UnitVec_k = \Cap_\NumData / \RangeKNtrue$, we can write:
\begin{equation}
\begin{aligned}
    \frac{\partial f_k}{\partial \ParamTrans}
    = 
    \left[
        \frac{\partial f_k}{\partial t_x} \;\;
        \frac{\partial f_k}{\partial t_y} \;\;
        \frac{\partial f_k}{\partial t_z}
    \right]^\top
    = 
    \frac{1}{\RangeKNtrue}
    \left[
        \rho_x \;\;
        \rho_y \;\;
        \rho_z
    \right]^\top
    = \UnitVec_k^\top.
\end{aligned}
\end{equation}

\subsection*{A.2) Derivative with respect to \texorpdfstring{$\ParamRotVec$}{TEXT}}

The derivative of $f_k$ with respect to $\ParamRotVec$ is:
\begin{equation} \label{eq:dfk_dv}
\begin{aligned}
    &\frac{\partial f_k}{\partial \ParamRotVec} 
    = \frac{1}{2 \RangeKNtrue} 
    \left(
    2\rho_x \frac{\partial \rho_x}{\partial \ParamRotVec}
    + 2\rho_y \frac{\partial \rho_y}{\partial \ParamRotVec}
    + 2\rho_z \frac{\partial \rho_z}{\partial \ParamRotVec}
    \right)  \\
\end{aligned}
\end{equation}

Next, we find $\frac{\partial \rho_x}{\partial \ParamRotVec} = \left[\frac{\partial \rho_x}{\partial \ParamRotVecxyz_x} \; \frac{\partial \rho_x}{\partial \ParamRotVecxyz_y} \; \frac{\partial \rho_x}{\partial \ParamRotVecxyz_z} \right]$.
Replace $\rho_x = \Cap_\NumData \cdot \UnitX$ and note that $\ParamTrans$, $\Wap$ and $\UnitX$ are constant vectors, we can rewrite $\partial \rho_x / \partial \ParamRotVecxyz_x$ as:
\begin{equation} \label{eq:rhox_vx}
\begin{aligned}
    \frac{\partial \rho_x}{\partial \ParamRotVecxyz_x}
    &= \frac{\partial (\Cap_\NumData \cdot \UnitX)}{\partial \ParamRotVecxyz_x} 
    = \frac{\partial \Cap_\NumData}{\partial \ParamRotVecxyz_x} \cdot\UnitX 
    +
    \frac{\partial\UnitX}{\partial \ParamRotVecxyz_x} \cdot
    \Cap_\NumData \\
    &= \frac{\partial (\ParamTrans + \ParamScale \AlignedPos_\NumData - \Wap)}{\partial \ParamRotVecxyz_x} \cdot \UnitX 
    + \mathbf{0} \cdot \Cap_\NumData  \\
    &= \frac{\partial (\ParamScale \AlignedPos_\NumData)}{\partial \ParamRotVecxyz_x} \cdot \UnitX
\end{aligned}
\end{equation}

From Appendix 2 in \cite{gallego2014derivative3Drot}, we have a the following result:
\begin{equation} \label{eq:result2}
\begin{aligned}
    \frac{\partial (\ParamRotMat \mathbf{a})}{\partial \ParamRotVecxyz_i}
    =
    \frac{\partial (\ParamRotMat \mathbf{a})}{\partial \ParamRotVec} \; \UnitVec_i,
\end{aligned}
\end{equation}
where $\mathbf{a} \in \R^3$ is independent of $\ParamRotVec$ and $i=\{x,y,z\}$. Combining Eq. (\ref{eq:rhox_vx}) and (\ref{eq:result2}), we get
\begin{equation} 
\begin{aligned}
    \frac{\partial \rho_x}{\partial \ParamRotVecxyz_x}
    &= 
    \left(
    \frac{\partial (\ParamScale \AlignedPos_\NumData)}{\partial \ParamRotVec} \; \UnitX
    \right)
    \cdot \UnitX
    = 
    \UnitX^\top \;
    \frac{\partial (\ParamScale \AlignedPos_\NumData)}{\partial \ParamRotVec} \; \UnitX
    = \UnitX^\top \PartialRv \UnitX
\end{aligned}
\end{equation}
where $\PartialRv = \frac{\partial (\ParamScale \AlignedPos_\NumData)}{\partial \ParamRotVec} \in \R^{3\times3}$. Notice that $\UnitVec_i^\top \PartialRv \UnitVec_j = \PartialRv_{i,j}$, i.e. extracting the $i$-th row and $j$-th column of $\PartialRv$. Hence
\begin{equation} 
\begin{aligned}
    \frac{\partial \rho_x}{\partial \ParamRotVecxyz_x}
    = \PartialRv_{1,1}.
\end{aligned}
\end{equation}
Similarly, we have
\begin{equation} 
\begin{aligned}
    &\frac{\partial \rho_x}{\partial \ParamRotVecxyz_y}
    = \UnitX^\top \PartialRv \UnitY
    = \PartialRv_{1,2}, \quad
    \frac{\partial \rho_x}{\partial \ParamRotVecxyz_z}
    = \UnitX^\top \PartialRv \UnitZ
    = \PartialRv_{1,3}, \\
    &\frac{\partial \rho_y}{\partial \ParamRotVecxyz_x}
    = \UnitY^\top \PartialRv \UnitX
    = \PartialRv_{2,1}, \quad
    \frac{\partial \rho_y}{\partial \ParamRotVecxyz_y}
    = \UnitY^\top \PartialRv \UnitY
    = \PartialRv_{2,2}, \\
    &\frac{\partial \rho_y}{\partial \ParamRotVecxyz_z}
    = \UnitY^\top \PartialRv \UnitZ
    = \PartialRv_{2,3}, \quad
    \frac{\partial \rho_z}{\partial \ParamRotVecxyz_x}
    = \UnitZ^\top \PartialRv \UnitX
    = \PartialRv_{3,1}, \\
    &\frac{\partial \rho_z}{\partial \ParamRotVecxyz_y}
    = \UnitZ^\top \PartialRv \UnitY
    = \PartialRv_{3,2}, \quad
    \frac{\partial \rho_z}{\partial \ParamRotVecxyz_z}
    = \UnitZ^\top \PartialRv \UnitZ
    = \PartialRv_{3,3}.
\end{aligned}
\end{equation}
As such:
\begin{equation} \label{eq:drxyz_dv}
\begin{aligned}
    &\frac{\partial \rho_x}{\partial \ParamRotVec}
    = 
    \begin{bmatrix}
        \frac{\partial \rho_x}{\partial \ParamRotVecxyz_x} & \frac{\partial \rho_x}{\partial \ParamRotVecxyz_y} & \frac{\partial \rho_x}{\partial \ParamRotVecxyz_z}
    \end{bmatrix}
    =
    \begin{bmatrix}
        \PartialRv_{1,1} & \PartialRv_{1,2} & \PartialRv_{1,3}
    \end{bmatrix}, \\
    &\frac{\partial \rho_y}{\partial \ParamRotVec}
    =
    \begin{bmatrix}
        \PartialRv_{2,1} & \PartialRv_{2,2} & \PartialRv_{2,3}
    \end{bmatrix}, \quad
    \frac{\partial \rho_z}{\partial \ParamRotVec}
    =
    \begin{bmatrix}
        \PartialRv_{3,1} & \PartialRv_{3,2} & \PartialRv_{3,3}
    \end{bmatrix}.
\end{aligned}
\end{equation}
Combining Eq. (\ref{eq:dfk_dv}) and (\ref{eq:drxyz_dv}), we get
\begin{equation} \label{eq:dfk_rhok}
\begin{aligned}
    \frac{\partial f_k}{\partial \ParamRotVec} 
    &= \frac{1}{\RangeKNtrue}
    \begin{bmatrix}
        \rho_x \PartialRv_{1,1} + \rho_y \PartialRv_{2,1} + \rho_z \PartialRv_{3,1} \\
        \rho_x \PartialRv_{1,2} + \rho_y \PartialRv_{2,2} + \rho_z \PartialRv_{3,2} \\
        \rho_x \PartialRv_{1,3} + \rho_y \PartialRv_{2,3} + \rho_z \PartialRv_{3,3} 
    \end{bmatrix}^\top
    = \frac{1}{\RangeKNtrue} \Cap_\NumData^\top \PartialRv \\
    &= \UnitVec_k^\top \PartialRv.
\end{aligned}
\end{equation}
Lastly, we use Result 1 in \cite{gallego2014derivative3Drot} to find $\PartialRv$. Result 1 states that
\begin{equation} \label{eq:dRo_dv}
\begin{aligned}
    &\frac{\partial (\AlignedPos_\NumData)}{\partial \ParamRotVec}
    = - \ParamRotMat \sksym{\VCp_k} 
    \frac{\ParamRotVec \ParamRotVec^\top + (\ParamRotMat^\top - \IdMat_{3\times3}) \sksym{\ParamRotVec}}{\norm{\ParamRotVec}^2} \\
    \Rightarrow \;\;
    &\PartialRv 
    = \frac{\partial (\ParamScale \AlignedPos_\NumData)}{\partial \ParamRotVec}
    = - \ParamScale \ParamRotMat \sksym{\VCp_k} 
    \frac{\ParamRotVec \ParamRotVec^\top + (\ParamRotMat^\top - \IdMat_{3\times3}) \sksym{\ParamRotVec}}{\norm{\ParamRotVec}^2}.
\end{aligned}
\end{equation}
Hence, $\frac{\partial f_k}{\partial \ParamRotVec} = \UnitVec_k^\top \PartialRv$ is fully defined.

\subsection*{A.3) Derivative with respect to \texorpdfstring{$\ParamScale$}{TEXT}}

The derivative of $f_k = \sqrt{\rho_x^2 + \rho_y^2 + \rho_z^2}$ with respect to $\ParamScale$ is:
\begin{equation} \label{eq:dfk_ds}
\begin{aligned}
    \frac{\partial f_k}{\partial \ParamScale} 
    &= \frac{1}{2 \RangeKNtrue} 
    \left(
    2\rho_x \frac{\partial \rho_x}{\partial \ParamScale}
    + 2\rho_y \frac{\partial \rho_y}{\partial \ParamScale}
    + 2\rho_z \frac{\partial \rho_z}{\partial \ParamScale}
    \right)  \\
    &= \frac{1}{\RangeKNtrue}
    \begin{bmatrix}
        \rho_x & \rho_y & \rho_z
    \end{bmatrix}
    \cdot
    \begin{bmatrix}
        \frac{\partial \rho_x}{\partial \ParamScale} & \frac{\partial \rho_y}{\partial \ParamScale} & \frac{\partial \rho_z}{\partial \ParamScale}
    \end{bmatrix} \\
    &= \frac{1}{\RangeKNtrue} \Cap_\NumData \cdot
    \frac{\partial \Cap_\NumData}{\partial \ParamScale}
    = \UnitVec_k^\top
    \frac{\partial (\ParamTrans + \ParamScale \AlignedPos_\NumData - \Wap)}{\partial \ParamScale} \\
    &= \UnitVec_k^\top \AlignedPos_\NumData.
\end{aligned}
\end{equation}
This completes the proof.

\end{appendices}

\balance
\bibliographystyle{IEEEtran}
\bibliography{IEEEabrv,./references}

\end{document}